# Revealing the Technology Development of Natural Language Processing: A Scientific Entity-Centric Perspective

Heng Zhang[1], Chengzhi Zhang[1, *], Yuzhuo Wang[2]

1. Department of Information Management, Nanjing University of Science and Technology, Nanjing, 210094, China
2. Department of Management Science and Engineering, School of Management, Anhui University, Hefei, 230601, China

**Abstract**: Most studies on technology development have been conducted from a thematic perspective, but the topics are coarse-grained and insufficient to accurately represent technology. The development of automatic entity recognition techniques makes it possible to extract technology-related entities on a large scale. Thus, we perform a more accurate analysis of technology development from an entity-centric perspective. To begin with, we extract technology-related entities such as methods, datasets, metrics, and tools in articles on Natural Language Processing (NLP), and we apply a semi-automatic approach to normalize the entities. Subsequently, we calculate the z-scores of entities based on their co-occurrence networks to measure their impact. We then analyze the development trends of new technologies in the NLP domain since the beginning of the 21st century. The findings of this paper include three aspects: Firstly, the continued increase in the average number of entities per paper implies a growing burden on researchers to acquire relevant technical background knowledge. However, the emergence of pre-trained language models has injected new vitality into the technological innovation of the NLP domain. Secondly, Methods dominate among the 179 high-impact entities. An analysis of the *z-score* trend about the top 10 entities reveals that pre-trained language models, exemplified by BERT and Transformer, have become mainstream in recent years. Unlike the trend of the other eight method entities, the impact of Wikipedia dataset and BLEU metric has continued to rise in the long term. Thirdly, in recent years, there has been a remarkable surge in popularity for new high-impact technologies than ever before, and their acceptance by researchers has accelerated at an unprecedented speed. Our study provides a new perspective on analyzing technology development in a specific domain. This work can help researchers grasp the current hot technologies in the field of NLP and contribute to the future development of NLP technologies. The source code and dataset for this paper can be accessed at https://github.com/ZH-heng/technology_development.

[*] Corresponding author: Chengzhi Zhang (zhangcz@njust.edu.cn).

## 1. Introduction

Technological advancement is an essential driver for the development of human society. Vaccines and antibiotics have led to a decrease in common infectious diseases. Advances in prevention and treatment techniques have also made a decline in cardiovascular disease morbidity and mortality (Dzau & Balatbat, 2018). The development of medical technology provides better protection for people's lives, but the lack of medical resources is still an urgent problem. The COVID-19 pandemic promotes the application of telemedicine technology to address this problem (Sigman, 2020). Mobile internet technology is one of the current hot technologies that is changing people's lives in every dimension. Such as agricultural production (Khan et al., 2022), health management (T. Liu & Liu, 2022), and remote learning (Hu, 2022). New technologies are constantly emerging today, but the development of new technologies cannot be separated from existing ones. For example, ChatGPT was launched in November 2022 by OpenAI. As one of the most popular Natural Language Processing (NLP) tools, it is fine-tuned from GPT3.5, a Generative Pre-trained Transformer model[1]. Based on the Transformer architecture, GPT has undergone four versions of development from GPT1.0 to GPT3.5. Therefore, exploring technology development enables us to understand the life cycle of technology from birth and application to decline. By learning from the successes and failures of past technologies, individuals can promote the advancement of new technologies in the future. Additionally, analyzing the technology development can provide insight into the development of various domains.

The existing analysis about development of disciplinary domains primarily employs research topic-based methods. Relevant studies have compiled the popular research topics in NLP for each year (Parmar et al., 2020), establishing topic taxonomy (Hingmire et al., 2021) and analyzing the temporal changes of key thematic terms (Mohammad, 2019). Research topics are usually generated from the text by unsupervised methods. However, obtaining normalized and accurate topics often requires manual modification of the results, which can be laborious. Additionally, research topics tend to be coarse-grained, focusing more on research questions or tasks. Technology is crucial for promoting the development of various fields, and NLP is a field that heavily relies on technological innovation. Therefore, our aim is to examine the development of NLP technologies at the micro-entity level. This will allow us to observe the development of the NLP domain from a different perspective than that of research topics. Specific types of entities can describe the technology more precisely than the topics. In this article, there are four types of entities associated with technology:

[1] https://openai.com/blog/chatgpt

method, dataset, metric, and tool, as illustrated in Table 1.

**Table 1. Sample sentences containing technology-related entities**

| No. | Sentences |
|---|---|
| 1 | First, we compare the LSTM *(Method)* encoder with the Transformer *(Method)* encoder. |
| 2 | As a seed corpus, we use the CNN/Dailymail news corpus *(Dataset)*. |
| 3 | The baseline combination weights were tuned to optimize BLEU *(Metric)*. |
| 4 | We use Stanford CoreNLP *(Tool)* to recognize named entities. |

Research methods are closely related to technology. Järvelin & Vakkari (1993) and Tuomaala et al. (2014) analyzed the development of the field of Library and Information Science by focusing on the development and changes of research methods. However, they used content analysis to manually annotate the categories of methods used in the papers, which was labor-intensive and time-consuming. The advancement of automatic entity recognition technology has enabled the extraction of scientific entities from academic literature on a large scale (C. Zhang et al., 2023). We can utilize this technology to analyze the technology development in NLP at a reduced cost. In this paper, we plan to investigate the development of technology in the NLP filed from the entity level. To do this, we collect the full-text data of academic papers in NLP, develop scientific entity recognition model to automatically extract technology-related entities, and construct entity co-occurrence networks to calculate the impact of entities and analyze the technology development in NLP. We are particularly interested in new technology-related entities in the NLP field since the 21st century. In this article, a new entity is defined as an entity that first appears in the NLP field, even though some entities may have existed in other fields before. Nonetheless, for the NLP field, these entities are still considered to be novel concepts before they were introduced to NLP. Specifically, we explore the following three research questions.

***RQ1*:** What is the development tendency in the number of NLP technology-related entities?

***RQ2*:** What new technologies have emerged since the 21st century that have driven the research development in NLP?

***RQ3*: How the popularity degree and speed of high-impact new technologies vary over time?**

Technologies in a domain do not develop uniformly due to limitations in theory, computer technology, and other factors. To measure the speed of technological development, we can examine the number of technology-related entities. Therefore, we established RQ1. We further set RQ2 to determine the new technologies that have propelled the advancement of NLP research in the past two decades, as well as the situation of high-impact new technology-related entities. Based on RQ2, we formulated RQ3 to examine how the popularity degree and speed of high-impact new technologies change over time. Rotolo et al. (2015) have proposed five characteristics of emerging technologies, two of which are prominent impact and relatively fast growth. Building upon this framework, this paper introduces two indicators, popularity degree and speed, to describe the impact and growth rate of the group of high-impact new technologies during different time periods. The former calculates the cumulative impact of top new entities over varying time periods, while the latter measures the average time it takes for them to achieve high impact.

The main contributions of this paper are reflected in the following three aspects.

Firstly, our proposal is to examine the technological development of a field by analyzing fine-grained entities, which has not been done before. Previous studies on development analysis have mainly used topic models. But the topics tend to have broad meanings. In this paper, we concentrate on the development of NLP technology by identifying four types of entities: methods, datasets, metrics, and tools.

Secondly, we utilize pre-trained language models to identify technology-related entities. Additionally, we have developed a semi-supervised data augmentation technique to increase the number of training samples and improve the model's robustness. This approach has been demonstrated to enhance entity recognition accuracy, achieving an $F_1$ score of 87.00 for the final model.

Thirdly, the analysis results indicate a recent flourishing of new technologies in the field of NLP, with an explosive growth in related entities. Additionally, high-impact new entities are more popular than ever before, and researchers are adopting emerging hot technologies at a faster pace.

All the data and source code of this paper are freely available at the GitHub website: https://github.com/ZH-heng/technology_development.

## 2. Related work

Productivity enhancements and social changes are often seen as a result of technological advancements. Analyzing the technological development in a particular field is a valuable endeavor. This study provides a novel analysis of technological development at the entity level. It uses the NLP domain as an example and integrates scientific entity recognition and entity co-occurrence networks. The works related to this study includes three parts: technology development analysis, scientific entity recognition and co-occurrence network of scientific entities.

### 2.1 Technology development analysis

Technology was defined as the knowledge of how to fulfill certain human purposes in a specifiable and reproducible way (Brooks, 1980). Skolnikoff (1993) also endorses this definition when discussing the relationship between “science” and “technology”. Furthermore, he points out that technology should include not only physical technologies (such as the printing press, the telescope, and the microscope) but also “social” technologies, such as codified systems of management or computer software. Mitcham (1994) summarized four aspects of technology in everyday life:

material objects (from kitchenware to computers), knowledge (including recipes, rules, and theories), activities (encompassing a series of processes including design, construction, and utilization), and volition (understanding the usage of technology and comprehending its consequences).

In scientific research, researchers are guided by questions and explore answers through effective approaches. Considering the definition of technology, the approaches to solving scientific problems can also be regarded as technology. Therefore, we define the technologies in the NLP field as the means by which researchers solve scientific problems or tasks. The process of solving NLP tasks involves the construction of datasets, the application of methods (such as algorithms and models), evolution with metrics, and the use and creation of tools. Thus, we define the four categories of entities, methods, datasets, metrics, and tools, as technology-related entities and extract and analyze these entities to reveal the development of technology in the NLP field.

Scientists analyze the development of technology from various angles in order to forecast future technologies. Coccia (2019) measured the dynamics of technological evolution for technological forecasting based on the theory of technological parasitism. Yun et al. (2022) discovered new technological opportunities through methods such as patent semantic analysis and sequence pattern mining. In recent decades, information technology has seen rapid growth, and the security-development process of computer-science technologies is also of great importance(Percia David et al., 2023). In addition, some methods for analyzing the evolution of topics are also worth considering, as the technology-related entities in this study can also be regarded as knowledge units that are more fine-grained than topics. Y. Zhang et al. (2017) visualized the evolutionary pathways of scientific topics through methods such as term identification and clustering. Huang et al. (2022)

constructed keyword networks and obtained word embeddings, then identified topics in the co-word networks using the community detection algorithm, and finally calculated the semantic similarity of topics between adjacent time periods to reveal their evolutionary relationships.

### 2.2 Scientific entity recognition

Named entity was initially defined as a “unique identifier” in the common domain, comprising three significant categories and seven minor categories: entities (organizations, persons, locations), times (dates, times), and quantities (monetary values, percentages) (Sundheim, 1995). Yet, in specific disciplines, the identifications of scientific entity categories have evolved. In the medical field, researchers typically extract entities such as genes, chemicals, diseases, and proteins. However, NLP experts tend to concentrate on different entity types, such as method, metric, and dataset.

Entity recognition in the common field has achieved high accuracy. The models designed by researchers have achieved $F_1$ exceeding 0.9 on both the CoNLL2003 and Resume universal entity datasets (Geng et al., 2023). In certain domains, the complexity of academic texts and the need for domain-specific knowledge have made entity recognition significantly more difficult than that in general domains. Ma et al. (2023) achieved an $F_1$ value of only 0.636 for recognizing Operation and Effect entities when mining procedural scientific information in the field of AI. Currently, scientific entity recognition in specific fields has the following issues: (1) Large-scale pre-trained language models have significantly outperformed traditional machine learning models in the entity recognition task. However, researchers have found that transferring pre-trained models trained on the general corpus to specific domains leads to a decrease in the accuracy of entity recognition. (2) Compared to the common domain, the concept of domain-specific entities is difficult to define clearly, and entity types are more complex, which poses a severe challenge in entity boundary detection.

Researchers have undertaken specific efforts based on the above two questions. Table 2 presents a comprehensive list of their work and contributions.

**Table 2. Related works of scientific entity recognition**

| Article | Algorithm/Model | Contribution |
|---|---|---|
| Beltagy et al. (2019) | SciBERT | Fine-tuning BERT based on papers from computer science and broad biomedical domain. SciBERT improves the representation of background knowledge in related areas such as AI and NLP. |
| Z. Liu et al. (2020) | FinBERT | FinBERT is the first domain-specific BERT pre-trained on financial corpora, which can transfer financial domain knowledge to downstream text mining tasks. |
| Lee et al. (2020) | BioBERT | BioBERT is the first domain-specific BERT pre-trained on biomedical corpora, which shows better performance in tasks such as biological NER and RE. |
| Chalkidis et al. (2020) | LEGAL-BERT | Fine-tuned based on BERT through English legal text, LEGAL-BERT enhances the representation of legal domain knowledge. |
| Wei, Su, et al. (2020) | CASREL | Two sequences are used to record the start or end of entities, respectively. This enhances the representation of entity boundaries. |
| Wei, Hong, et al. (2020) | BiSELF-CRF | They employed a pointer network to reposition the boundary in aspect extraction, thus improving the accuracy of the extraction results. |

In this study, we aimed to address the problem of extracting multi-class entities in NLP papers. Therefore, based on SciBERT, we designed a scientific entity recognition model by combining the cascaded binary tagging framework proposed by Wei, Su, et al. We enhanced the representation of entity boundaries by using cascaded binary tagging and added a dedicated sequence to encode entity types.

### 2.3 Co-occurrence network of scientific entities

Network analysis is a widely used technique in fields such as bibliometrics and social network analysis. Researchers have also utilized dynamic networks to explore technological changes (Funk

& Owen-Smith, 2017). Network comprises two essential elements: the nodes and the weights of the edges between nodes. In prior studies, researchers constructed networks based on literature content for pertinent bibliometric analysis, and the network nodes included: authors, organizations, keywords, subject terms, words, etc. Methods such as co-occurrence frequency and similarity are then utilized to measure the weights between nodes. Li et al. (2016) construct articles co-keyword networks using articles as nodes, co-keyword relationships as edges, and the number of co-keywords as weights. They also built keyword co-occurrence networks with article keywords as nodes, co-occurrence relationships as edges, and the number of co-occurrences as weights. Mao et al. (2020) applied co-word networks to keyword extraction by using the co-occurrence frequency of words in a fixed window and the semantic similarity between words, etc., to calculate the weights between network nodes when constructing the co-word networks. The networks can be broadly classified into two types: directed (citation network) and undirected (word co-occurrence network). With the comprehensive examination of full-text content, researchers shifted their focus toward entities in academic literature and utilized them as nodes for constructing related research networks. Ding et al. (2013) proposed the concept of Entitymetrics, and they built entity citation networks based on the citation relationships of the literature to measure the impact of entities. Li & Yan (2018) extracted software entities from PLOS journal articles and constructed a co-mention network of R packages to discover the top packages mentioned in the papers.

Calculating the significance of network nodes is a crucial task in network analysis, and it involves various indexes such as Centrality, PageRank, *z-score*, and so on. The first two do not take into account the weights between nodes, while the optimized *z-score* considers the weights between nodes when calculating the importance of nodes. Guimerà et al. (2007) proposed calculating the *z-* *score* in the network module to measure how "well-connected" node i is with other nodes. But the weights between nodes are not considered. Wang et al. (2014) used *z-score* to rank words within the network community when analyzing the evolution of the research topic. Huang et al. (2022) also utilized the *z-score* to rank the nodes within the community but calculated it by adding the weights between nodes.

The research of Yao et al. (2023) is similar to ours. We both explore the development of a specific domain from an entity's perspective. It is noted that we focus on the new technology-related entities that have led the development of the NLP field since the 21st century. By analyzing the changes in highly-impact new entities, we can reveal the development of NLP technologies. Furthermore, this study differs from theirs in the following aspects. The field we study is NLP, while they explore AI. We calculate the *z-score* by constructing co-occurrence networks to measure the impact of entities rather than simply counting the frequency. We develop a semi-automatic entity normalization method, whereas they rely entirely on manual entity normalization.

## 3. Methodology

The objective of our study is to investigate the development of technology in the filed of NLP from an entity perspective. To achieve this, we utilized an automated scientific entity recognition method to extract entities related to technology from NLP papers. We then built entity co-occurrence networks and computed *z-values* to evaluate the impact of these entities. Figure 1 illustrates the framework of this study. Specifically, we accomplish this study in four steps. The first step involves constructing the dataset. To do so, we gathered PDF papers in the field of NLP, parsed their contents, and randomly selected a subset of articles to annotate technology-related entities. This allowed us to build training datasets for our study. In the second step, we identified the entities in all papers. To accomplish this, we trained NER models based on annotated datasets and designed a semi-supervised data augmentation method to enhance the model's performance and robustness. By evaluating the results, we identified the best model for extracting technology-related entities from full-text of papers. The third step involves *z-score* computing. First, we normalized all identified entities, and then constructed their co-occurrence networks based on their co-mentions in papers. After that, we computed z-scores of entities in co-occurrence networks to measure their impact. The fourth step is to analyze the development of technology-related entities. We analyzed the annual changes in the number of entities, the situation of high-impact new entities, and the variations in the popularity degree and speed of top entities.

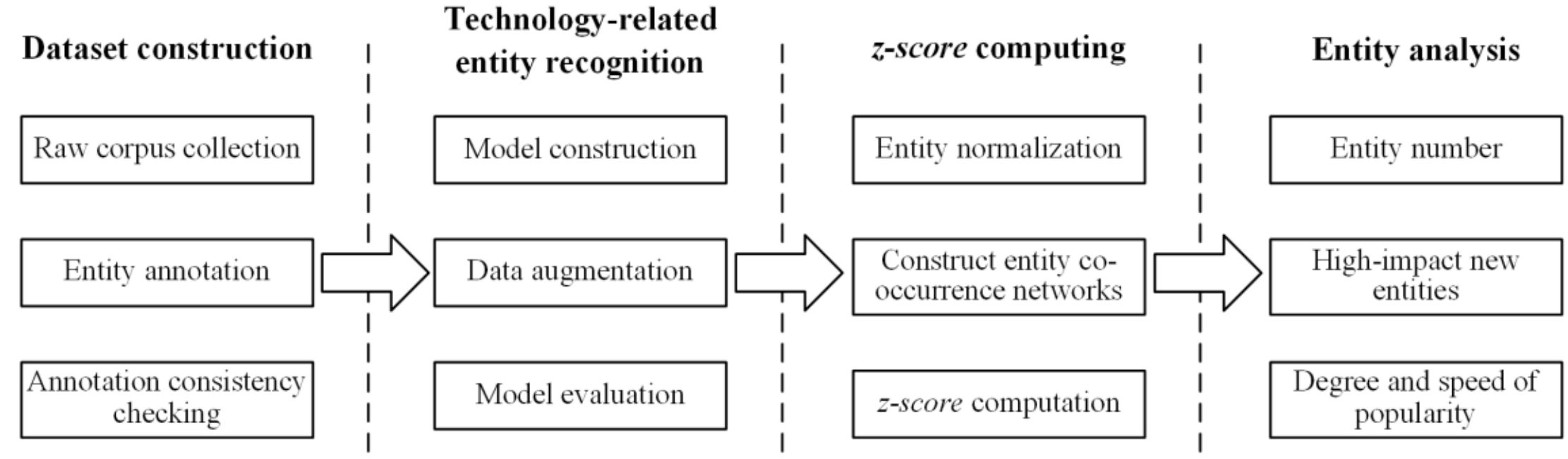

**Figure 1. Research framework of technology development analysis**

### 3.1 Dataset construction

#### (1) Raw corpus collection

Most conference proceedings in NLP are available on the ACL Anthology website[2]. We selected three representative conferences for our study: ACL (Annual Meeting of the Association for Computational Linguistics), EMNLP (Conference on Empirical Methods in Natural Language Processing), and NAACL (North American Chapter of the Association for Computational Linguistics). And we downloaded all their PDF papers from 2000 to 2022. Then, we utilized the open-source tool pdf2htmlEX[3] to convert these papers into HTML format and extracted their contents using regular expression. A total of 17,783 articles were successfully converted. Figure 2 illustrates the paper numbers for each year of the three conferences. It shows a yearly increase in the number of papers in the NLP field. The ACL and EMNLP conferences are held annually, but the NAACL conference is not held every year, resulting in a zero count of corresponding papers in seven years.

[2] https://aclanthology.org/
[3] https://pdf2htmlex.github.io/pdf2htmlEX/

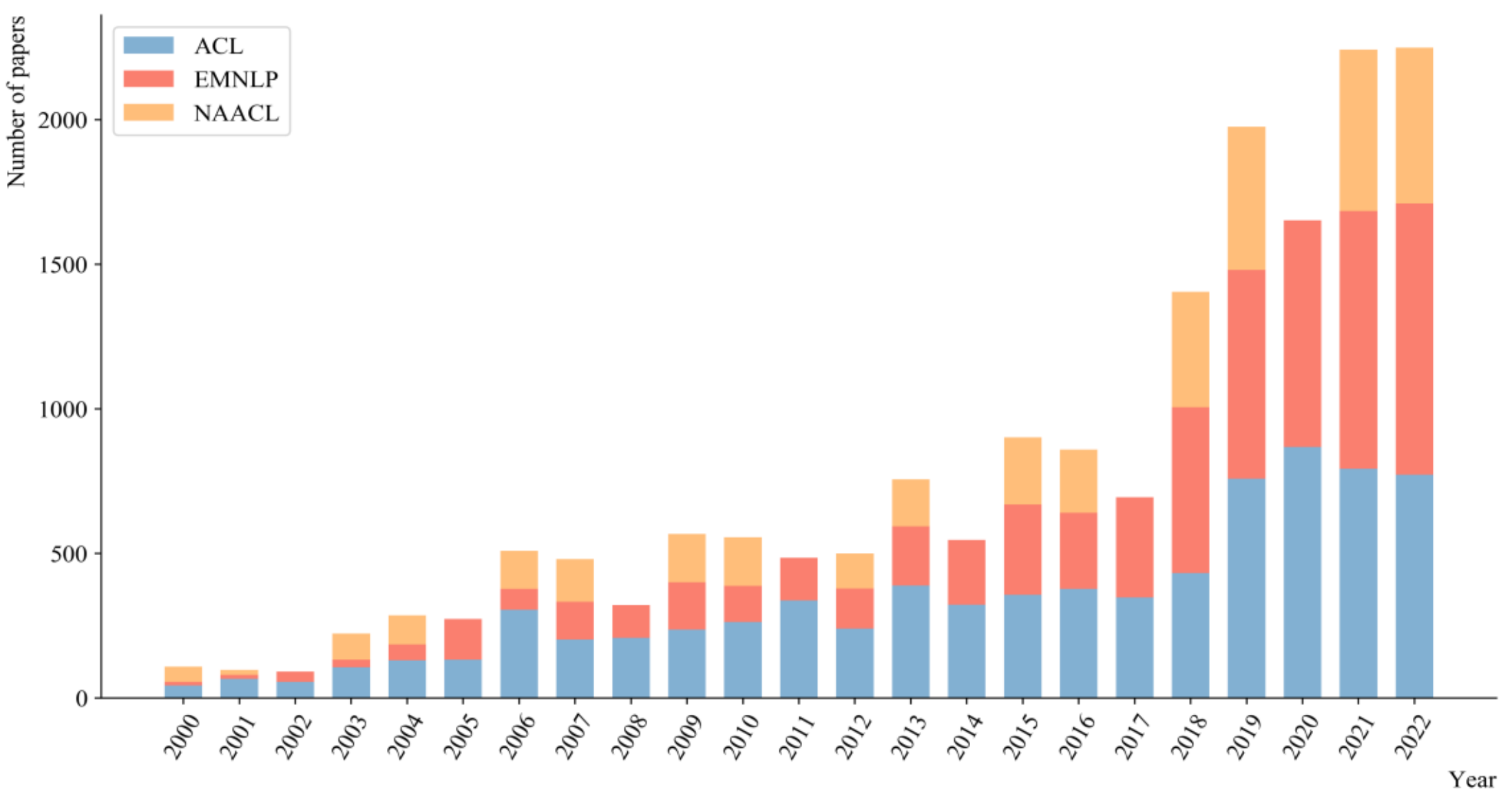

**Figure 2. Number of papers for the three conferences in each year**

**(2) Entity annotation**

Two entity datasets related to the field we studied are currently available: SciERC and TDM. The SciERC dataset is constructed from conference papers in AI, NLP, machine learning, etc. The dataset consists of six entity types, namely, Task, Method, Metric, Material, Other-ScientificTerm, and Generic (Luan et al., 2018). The TDM dataset is built from NLP articles, defining three types of entities: Task, Dataset, and Metric (Hou et al., 2021). This paper concentrates on technology-related entities and describes four types: method, dataset, metric, and tool. Their descriptions and examples are provided in Table 3. Given the variations in entity types between SciERC, TDM datasets, and the specific entities of interest to us, as well as our observation that abstracts may not fully contain all four types of technology-related entities we need, we have made the decision to not directly utilize these two datasets. Therefore, we randomly selected 50 full-text ACL papers for entity annotation to create our own dataset.

**Table 3. The descriptions and examples of four entity types**

| No. | Type | Descriptions | Examples |
|---|---|---|---|
| 1 | Method | Algorithms, models, etc. | *SVM, LSTM, BERT, Adam, RNN* |
| 2 | Dataset | Corpus, dataset, dictionary, etc. | *Brown Corpus, Penn Treebank, WordNet* |
| 3 | Metric | Indictors which are used to evaluate experimental results. | *Accuracy, Precision, Recall, $F_1$-score, BLEU* |
| 4 | Tool | Programming languages, software, open-source tools, etc. | *Python, GIZA++, TensorFlow, PyTorch* |

Our entity annotation process is as follows:

*Pre-annotation and refine annotation guidelines.* The annotation work was carried out by five NLP researchers using the WebAnno[4] annotation tool. Initially, two annotators pre-annotated the full text of 14 papers. They resolved conflicting results and refined the annotation guidelines. Table 3 provides a concise overview of the annotation guidelines.

*Follow the guidelines to annotate entities.* The other three annotators followed the annotation guidelines to annotate the entities in the remaining 36 papers. To guarantee the quality of annotations, we split 36 papers into three groups, each comprising 12 papers. We formed three teams by pairing up three annotators, with every team assigned to annotate one group of papers. This approach guarantees that two different annotators will annotate each paper and that every annotator will have annotated 24 papers.

*Check the consistency of annotations*. We calculated Cohen's Kappa (McHugh, 2012) values between two pairs of the three annotators in WebAnno. The resulting values were 0.90, 0.91, and 0.83, respectively, indicating a strong level of agreement among annotators. Therefore, our annotation dataset is reliable for training the entity recognition model.

**(3) Statistics on the dataset we annotated**

[4] https://webanno.github.io/webanno/

We have annotated entities of 2815 sentences and conducted statistics on various entities, as shown in Table 4. We observed that the data distribution is unbalanced, and the number of tool entities is significantly smaller than the other three types. Therefore, we will use this dataset to train a model and extract entities from unannotated texts. We reviewed the sentences containing the model's predicted entities and selected 300 sentences to add to the original 2815 sentences, forming the entity recognition dataset. Therefore, our dataset contains a total of 3115 sentences, while SciERC and TDM contain 2687 and 2010 sentences respectively.

**Table 4. Statistical information of the annotation dataset**

| Type | #Sentences | #Entities |
|---|---|---|
| Method | 1794 | 2546 |
| Dataset | 591 | 840 |
| Metric | 609 | 896 |
| Tool | 142 | 155 |

We then split all 3115 sentences into training set (Trainset), validation set (Validset), and testing set (Testset) using an 8:1:1 ratio. The number of various entities in each set was adjusted by setting random seeds to ensure data balance. The statistical results of the training, validation, and testing sets are presented in Table 5.

**Table 5. Statistical results of Trainset, Validset, and Testset**

| Dataset | #Sentences | #Entities | | | |
|---|---|---|---|---|---|
| | | Method | Dataset | Metric | Tool |
| Trainset | 2493 | 2180 | 700 | 753 | 445 |
| Validset | 311 | 286 | 85 | 77 | 65 |
| Testset | 311 | 268 | 85 | 79 | 44 |

### 3.2 Entity recognition model and evaluation

We constructed a technology-related entity recognition model based on SciBERT and the cascaded binary tagging framework, as shown in Figure 3. We use two sequences to represent the start and end positions of entities. The sequences use 0/1 to indicate whether a token is the start or end of the entity. Another sequence is utilized to indicate the entity type.

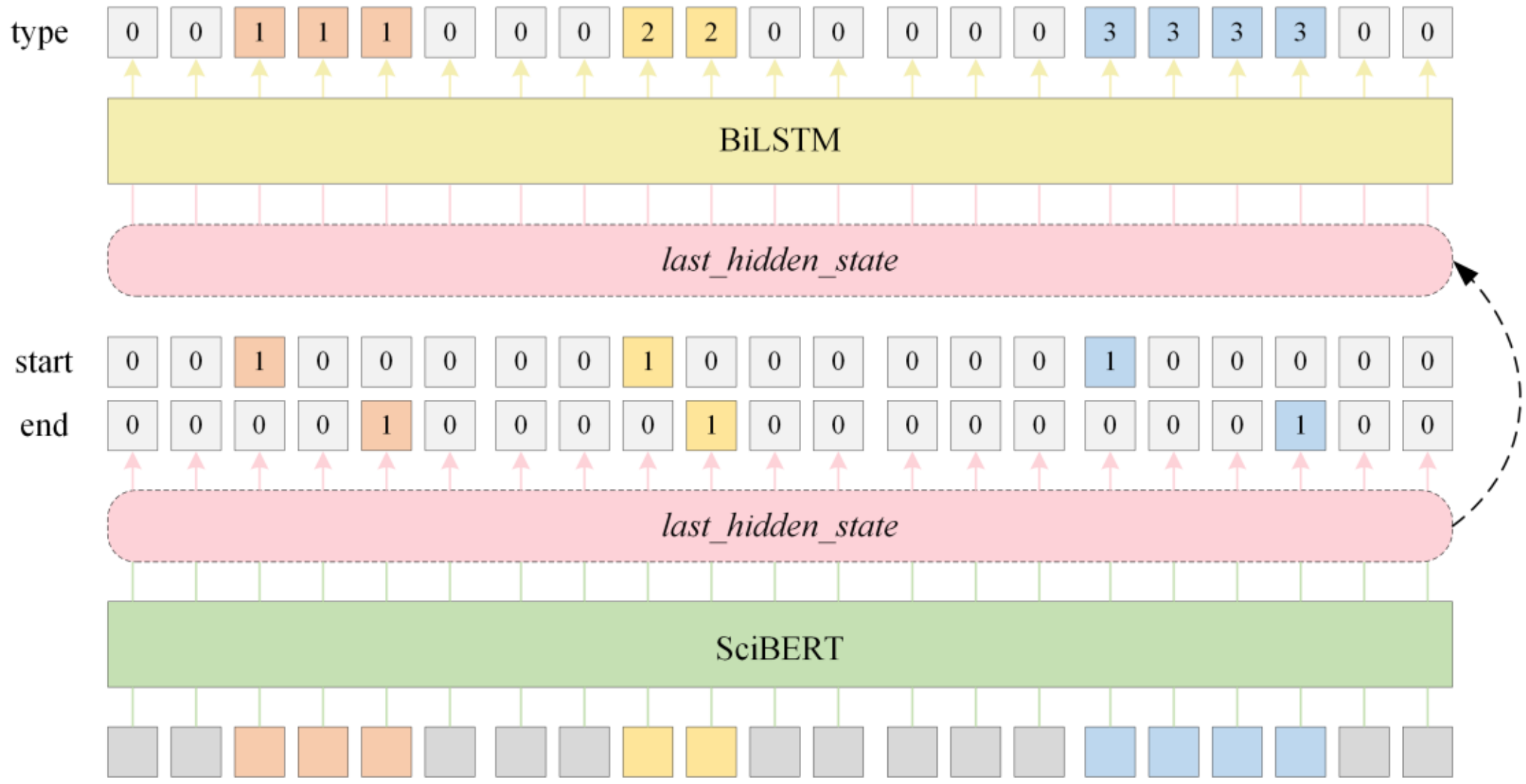

(The cascade binary tagging framework was enhanced in this study to accommodate the multi-class entity extraction task. Specifically, two binary sequences were utilized to encode the start and end positions of the entity, while another sequence was employed to encode the entity type.)

**Figure 3. Enhanced cascade tagging NER model for scientific entity extraction**

The input $X$ of the model is an $m*n$ matrix, $X = \{x_i\}_{i=1}^{m} = \{x_1, x_2, \ldots, x_m\}$, and $x_i = (x_{i2}, x_{i2}, \ldots, x_{in})$. We feed the data into the model for training in batches, and $m$ is batch size. $x_i$ represents a tokenized sentence, where each element represents a single token. In each batch of data, we take the length of the longest sentence as the value of max_len, and $n$ is equal to max_len.

Once $X$ is input into the model, the pre-trained SciBERT utilizes domain background knowledge to encode the text. The last hidden state of SciBERT is used as the text representation, denoted as $h$.

$$h = SciBERT(X) \tag{1}$$

Next, fully connected layers are created to map separately onto the start and end labels, obtaining the corresponding logits.

$$logits_{start} = h * W.T + b \tag{2}$$

$$logits_{end} = h * W.T + b \tag{3}$$

Subsequently, the CrossEntropyLoss function is used to calculate the loss value, where $l_{start}$ and $l_{end}$ are used to respectively represent the encoded start and end labels.

$$Loss_{start} = -\sum_j logits_{start}(j) * \log(l_{start}(j)) \tag{4}$$

$$Loss_{end} = -\sum_j logits_{end}(j) * \log(l_{end}(j)) \tag{5}$$

On the other hand, the text representation $h$ is fed into a BiLSTM network to further acquire contextual information.

$$h' = BiLSTM(h) \tag{6}$$

Then, we construct a fully connected network to map onto the category labels ($l_{type}$), and also use the CrossEntropyLoss function to calculate the loss value.

$$logits_{type} = h' * W.T + b \tag{7}$$

$$Loss_{type} = -\sum_j logits_{type}(j) * \log(l_{type}(j)) \tag{8}$$

Finally, the loss of model is the sum of the three losses.

$$Loss = Loss_{start} + Loss_{end} + Loss_{type} \tag{9}$$

In addition, we constructed baseline models utilizing BERT, RoBERTa, SciBERT, and T5. The inputs were encoded using BIOES tags. BERT is a self-encoding model based on bidirectional Transformers. It was introduced in 2018 by four researchers from Google AI Language and is considered a significant milestone in pre-trained language models. RoBERTa has optimized training data, time, methods, and other aspects based on BERT. SciBERT and BERT share the same model structure. However, SciBERT's training corpus comprises 1.14 million scientific articles from various domains, including biomedical, computer science, and natural language processing. This enables SciBERT to acquire background knowledge in relevant fields and enhance its capacity to represent the semantics of scientific texts. The T5 model adopts the Transformer architecture and can handle various natural language processing tasks. It transforms all NLP tasks into a text-to-text format and uses a unified network structure.

We evaluated the performance of different models using accuracy (P), recall (R), and micro-$F_1$ metrics. And we selected the optimal model for scientific entity identification of unannotated articles.

**3.3 Data augmentation**

Due to the exorbitant cost associated with data annotation, insufficient training data is often encountered in NLP tasks. To construct a dataset specifically for scientific entity recognition in this paper, several months were dedicated to obtaining a modest 3115 annotated sentences. In comparison, two open datasets, SciERC and TDM, possess a smaller scale. Training models on such limited data does not ensure optimal performance and robustness. To address this issue, researchers have proposed various text data augmentation techniques that leverage existing annotated data to expand the training samples. For instance, employing easy data augmentation methods, such as Synonym Replacement, Random Insertion, Random Swap, and Random Deletion, enriches the training samples and enhances the performance of text classification tasks (J. Wei & Zou, 2019).

However, as entity recognition data annotation involves marking specific words or phrases within sentences, directly using methods like easy data augmentation is not convenient. This approach may change the entities within sentences and render the annotated samples unusable. Therefore, we have designed a semi-supervised method to expand the training samples.

To ensure data diversity, we extracted all annotated entities from the training set and all the unannotated abstracts. Then, we matched sentences containing entities from the abstracts as candidate samples. However, there is a risk of "insufficient annotating" when directly matching, where entities that are not included in the entity set cannot be matched, or only a portion of the complete entity can be matched. Therefore, we built a model based on the training set to identify entities in the candidate samples. We only keep the sentences if all matched entities appear in the identified entities. By further filtering candidate samples in this way, we can solve the "insufficient annotating" problem to a certain extent. After this, we selected the same number of sentences as in the training set from the filtered candidate samples and added them to the training set for model training.

### 3.4 Semi-automatic entity normalization

Entity normalization, also known as entity disambiguation, is a common issue encountered with scientific entities extracted from academic papers. This problem is akin to disambiguating people names, institutions, and words. Diversity in these entities is expressed in different ways, with variations in the expression of the same entity in various papers. Ambiguity, where entities with the same term have distinct meanings in different contexts, is also present, as the same abbreviation may correspond to several separate entities.

In this paper, we normalize scientific entities in two stages. Firstly, we convert shortened entities with high frequency to full name format. Then, we calculate similarities between entities based on their word characteristics and normalize the entities using a clustering method. We describe the steps involved in detail below.

**(1) Constructing an "abbreviation-full name" mapping dictionary for entities**

We extract the entities containing brackets from the identification results and split them based on the brackets. The shorter part is considered the entity abbreviation, while the longer part is the full name. We obtained a mapping dictionary of high-frequency "abbreviation-full name" pairs through manual auditing. Examples of this are listed in Table 6, where the same abbreviation may correspond to multiple different full names.

**Table 6. Examples of "abbreviation-full name" dictionary**

| Abbreviation | Full name |
| --- | --- |
| CNN | Convolutional Neural Network |
| MRR | Mean Reciprocal Rank; Mean Replacement Rate |
| LSTM | Long Short Term Memory |
| SVM | Support Vector Machine |
| AP | Average Precision; Average Proportion; Affinity Propagation |
| CRF | Conditional Random Field |
| CRFs | Conditional Random Field |

**(2) Replacing the abbreviation with the full name**

We assume that an abbreviated entity expresses only one meaning within the same paper, so we take the paper as a unit to identify the corresponding full name of the abbreviation and replace it accordingly. we adopted the fastText[5] to train a word embedding model based on the collected full texts of papers to generate representation vectors for each entity. If an abbreviated entity from the dictionary appears in the paper, we calculate the similarity of each full name corresponding to that abbreviated entity to all entities in the article. The full name with the highest similarity is considered as the full name of the abbreviated entity, and the identified full name is used to replace the abbreviated entity in the text.

**(3) Calculating the edit distance similarity between all full-name entities**

After replacing the abbreviations of entities in the paper with their full names, we also perform lemmatization on the entities. Then, words with little semantic contribution (e.g., model, method, etc.) are removed, and spaces within entities are deleted. Finally, the edit distance similarities between the processed entities are computed.

**(4) Hierarchical clustering-based entity normalization**

The calculation of hierarchical clustering can be time-consuming when dealing with a large number of entities. To address this issue, we propose a simplified approach. We first initialize each entity pair with a similarity above 0.95, as well as each individual entity, as a separate cluster. To determine the similarity between two clusters, we calculate the average similarity of all entity pairs across them. In each iteration, we merge two clusters when their similarity exceeds the threshold of 0.85 (After observing the edit distance similarity of specific entity pairs and the clustering results under different thresholds, we ultimately determined the optimal threshold), rather than merging only the two clusters with the highest similarity. We repeat this process until the number of clusters no longer changes.

[5] https://github.com/facebookresearch/fastText/

Each cluster in the clustering result represents one entity. Furthermore, we consider that if a new technology is introduced in the NLP field for the first time in a paper, the corresponding entities will certainly be mentioned multiple times in the paper. Therefore, we annually calculate the total frequency of the original entities corresponding to each entity cluster. If the frequency of an entity cluster is less than 5 each year, we filter it out since it may be an entity misidentified by the model. With the above steps, we achieve scientific entity normalization in a semi-automatic way.

### 3.5 *z-score* computing of scientific entities based on co-occurrence network

Many existing studies have counted frequencies to measure impact. Since researchers often use a combination of methods to address research questions, we calculate the *z-scores* of technology-related entities in the co-occurrence network to measure their impact. We construct entity co-occurrence networks based on the normalized technology-related entities. The normalized entities are used as nodes in the network, the co-occurrence relationships of entities in the same paper form edges, and the number of papers mentioning both entities is taken as the weight of the edge between them. Furthermore, in order to examine the changes in the impact of entities in different years, we construct entity co-occurrence networks based on the paper data of each year and the entities extracted from them separately. The identification of entity co-occurrence relationships and the statistics of edge weights were both limited in the same year. The technology-related entities co-occurrence network of 2022 is shown in Figure 4.

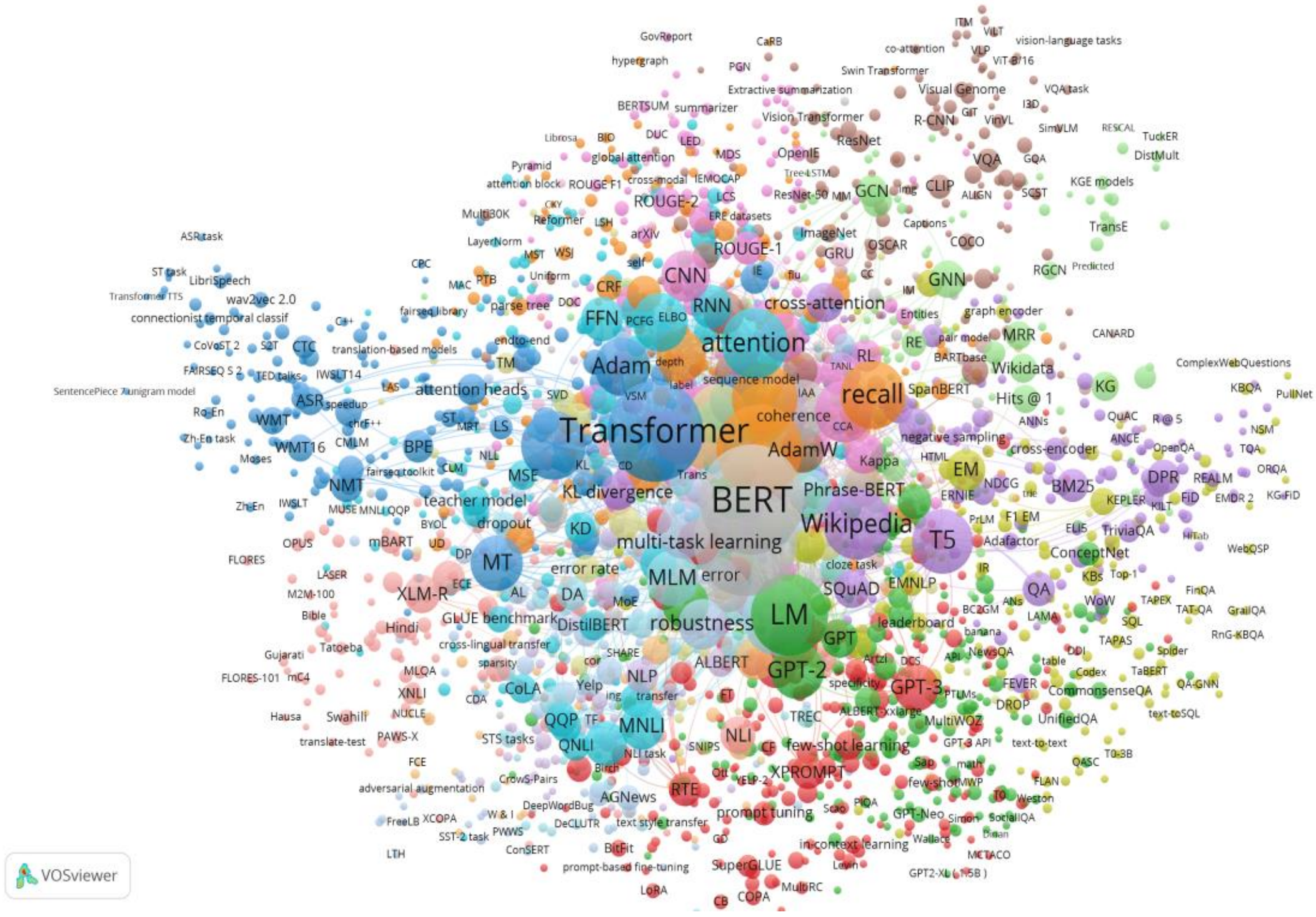

**Figure 4. Technology-related entities co-occurrence network of 2022**

Then, the *z-scores* of entities are calculated based on entity co-occurrence networks as follows.

$$z - score_i = \frac{W_C^i - \frac{1}{N}\sum_{j\in C} W_C^j}{\sqrt{\frac{1}{N}\sum_{j\in C}(W_C^j)^2 - (\frac{1}{N}\sum_{j\in C} W_C^j)^2}} \tag{10}$$

Where, $W_C^i$ : sum of edges weights of entity $i$ with other entities in network community $C$. $N$: number of entities in co-occurrence network. The higher the *z-score* of an entity indicates, the closer it is to other entities and the more significant its impact.

**4. Results**

In this section, we present and analyze the results of technology-related entity identification from NLP papers. And we discuss three research questions combined with the *z-score* calculation of entities.

### 4.1 Technology-related entity identification

In this article, we measure the performance of models using Precision (P), Recall (R), and $F_1$. We train the models on the Trainset, keep the best-performing checkpoint on the Validset for predicting the entities in the Testset, and compute the Precision, Recall, and $F_1$ scores with the annotated entities. We compare the results of our designed model with several baseline models on our annotated NLP corpus. Additionally, we compare our model with existing works on two relevant open datasets, namely SciERC and TDM.

**(1) Model evaluation on our annotated dataset**

We have designed a cascade model architecture with SciBERT+BiLSTM and applied a semi-supervised data augmentation method (data_aug). Furthermore, we have constructed four baseline models based on BERT, RoBERTa, SciBERT, and T5, and then built another four baseline models by adding a CRF layer to each of them. As indicated in Table 7, the scientific entity recognition results of the pre-trained model SciBERT based on the relevant domain corpus are significantly superior to other pre-trained models trained on the common corpus. The cascaded SciBERT + BiLSTM model that we devised attains an $F_1$ score of 86.49, surpassing all baseline models. Incorporating data augmentation techniques lead to a further improvement in the $F_1$ score to 87.00. The cascade model structure avoids the issue of too many entity labels when recognizing of multi-type entities. After the sentences are encoded by the pre-trained model, the hidden layer is directly used to judge whether a token is the start or end of an entity. Meanwhile, we fully connect the hidden layer with Bi-LSTM to identify the type of entities using contextual information. The application of semi-supervised data augmentation technique increases the diversity of training samples and enables the model to learn more contextual information, resulting in a further enhancement in performance.

Hence, our model achieves an efficient entity recognition result.

**Table 7. Evaluation of models on our annotated dataset**

| Model | P | R | $F_1$ |
|---|---|---|---|
| BERT | 83.64 | 85.92 | 84.77 |
| BERT+CRF | 84.10 | 84.45 | 84.28 |
| RoBERTa | 85.84 | 81.51 | 83.62 |
| RoBERTa+CRF | 83.86 | 82.98 | 83.42 |
| T5 | 84.41 | 85.29 | 84.85 |
| T5+ CRF | 85.08 | 85.08 | 85.08 |
| SciBERT | 85.95 | 86.13 | 86.04 |
| SciBERT+CRF | 83.53 | 87.39 | 85.42 |
| SciBERT+BiLSTM (cascade) | 86.22 | 86.76 | 86.49 |
| SciBERT+BiLSTM (cascade)+data_aug | 86.82 | 87.18 | **87.00** |

**(2) Evaluation on SciERC and TDM**

To further evaluate the entity recognition performance of our model on other datasets. We collected two relevant open datasets, namely SciERC and TDM, to compare the evaluation results of our model with existing models. Table 8 shows that our model achieved the highest $F_1$ score of 69.77 on the TDM dataset and performed well on the SciERC dataset. The result demonstrates the excellent generalization ability of our method for scientific entity recognition in the NLP domain.

**Table 8. Evaluation on SciERC and TDM dataset**

| Dataset | Authors | Model | P | R | $F_1$ |
|---|---|---|---|---|---|
| SciERC | Luan et al. (2018) | SCIIE | 67.20 | 61.50 | 64.20 |
| | Zhong and Chen (2021) | PURE | | | 68.90 |
| | Eberts and Ulges (2021) | SpERT | 70.87 | 69.79 | 70.33 |
| | Zaratiana et al. (2022) | Hierarchical Transformer | 67.99 | 74.11 | **70.91** |
| | Our | SciBERT+BiLSTM (cascade)+data_aug | 66.95 | 71.49 | 69.14 |
| TDM | Hou et al. (2021) | SCIIE | 67.17 | 58.27 | 62.40 |
| | Zaratiana et al. (2022) | Hierarchical Transformer | 65.56 | 70.21 | 67.81 |
| | Our | SciBERT+BiLSTM (cascade)+data_aug | 68.84 | 70.73 | **69.77** |

**(3) Experimental setup**

All models are trained and tested on the A5000 GPU. And we adjusted appropriate parameters for different models to achieve their respective optimal performance.

### 4.2 Number growth the number of technology-related entities in NLP

In this section, we answer RQ1. We utilized the SciBERT+BiLSTM cascade model and semi-supervised data augmentation to automatically identify technology-related entities in NLP papers, and a total of 534,500 entities were extracted, and the number of entities after normalization was 268,392. Subsequently, we filtered out entities with an annual frequency of less than 5, and ultimately obtained 37,624 valid technology-related entities. Each valid entity corresponds to a cluster in the clustering result. To verify the effectiveness of entity normalization, we randomly selected 1000 pairs of entities from these entity clusters and manually judged whether they belonged to the same entity. The precision metric, calculated based on the human reviews and the normalization results, was 91.30. Next, we analyzed the development tend of technologies in NLP from the average number of technology-related entities contained in a single paper, the growth of new entities, and so on.

**(1) The average number of entities contained in a single paper is increasing year by year**

We counted the non-repeating entities in each paper and calculated the mean quantity of technology-related entities included in articles for each year, which is plotted in Figure 5.

Overall, the average number of entities has been continuously increasing. In 2000, each article contained an average of 7.7 technology-related entities, which has risen to 45.1 in 2022, representing a growth of nearly fivefold! Among the 37,624 entities obtained after normalization and filtering, method entities account for 71.22%. This is significantly higher than the other three types of entities, and the growth trend of method entities is almost identical to that of the overall entities. Each paper's average number of method entities increases from 3.7 in 2000 to 29.3 in 2022. The growth curve for dataset, metric, and tool entities shows a relatively smooth and minor increase. The increasing average number of technology-related entities in each paper implies that researchers need to acquire more relevant background knowledge about technologies and learn more research methods. Consequently, research works in NLP are becoming more and more challenging.

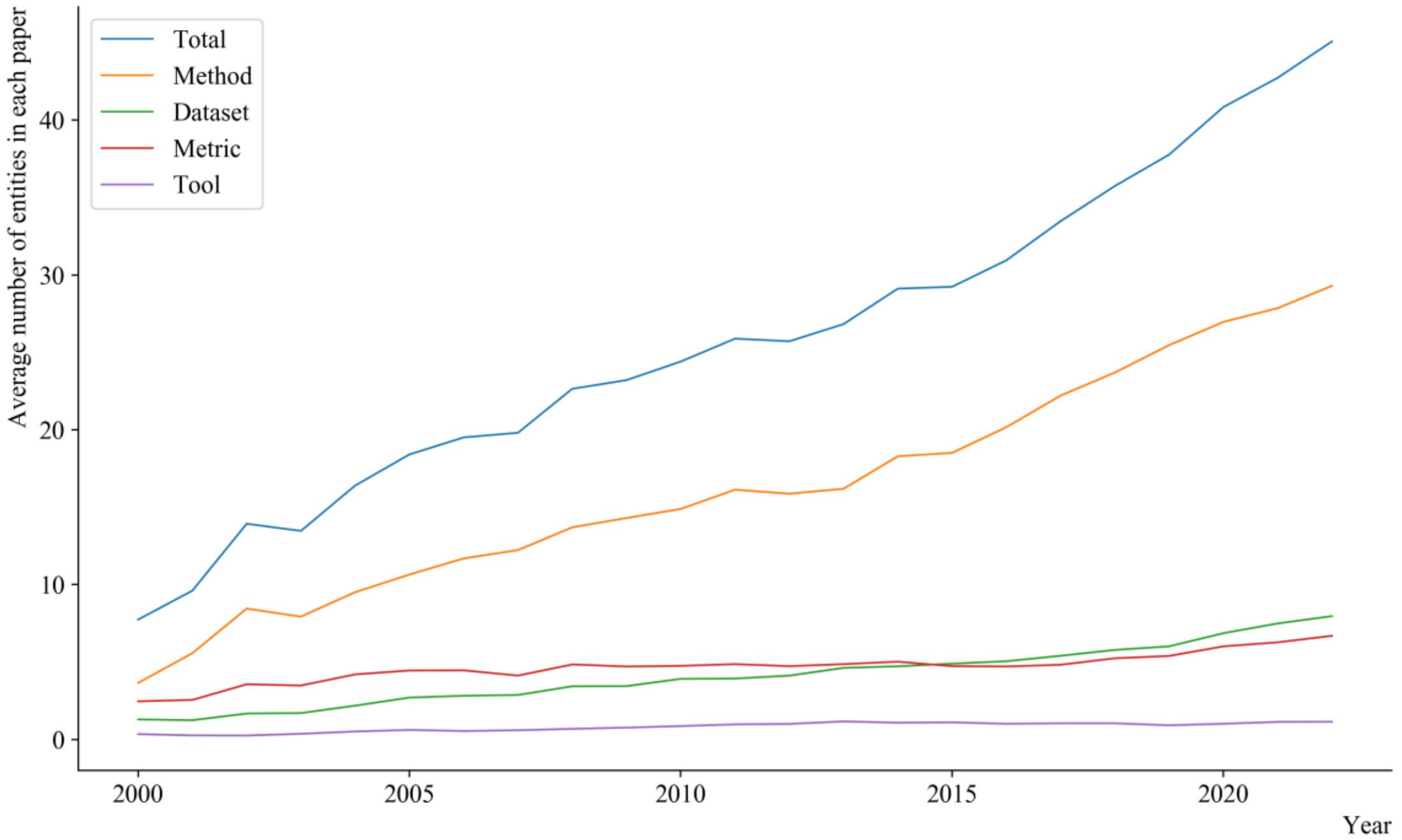

**Figure 5. The average number of technology-related entities included in one article since 2000**

**(2) New technology-related entities explode in the era of pre-trained language models**

Errors in entity recognition and normalization may lead to mistakes in determining the year when an entity first appears. Considering that when researchers first introduce a new technology into the field of NLP, they are likely to mention the corresponding entities multiple times in their papers, we take the year in which an entity appears with a first frequency of 5 or more as the year of its initial appearance. As our collected data on earliest papers only dated back to 2000, it was not possible to determine whether the entities appearing in that year were new or had already existed. As a result, we began counting the number of new entities in NLP from 2001 to 2022, as depicted in Figure 6. To eliminate the impact of the increasing number of research papers, we calculated the average number of new entities by dividing the total number of new entities each year by the number of papers published in that year. We depicted this information in the form of a bar graph on Figure 6, with the values aligned with the scale on the right y-axis. Each year, researchers in the NLP domain propose a variety of new technology-related entities. Although the overall number of new entities is increasing across all four types, the growth rate varies for each type. Specifically, method entities are the most numerous among the new entities, followed by dataset, with fewer metric and tool entities.

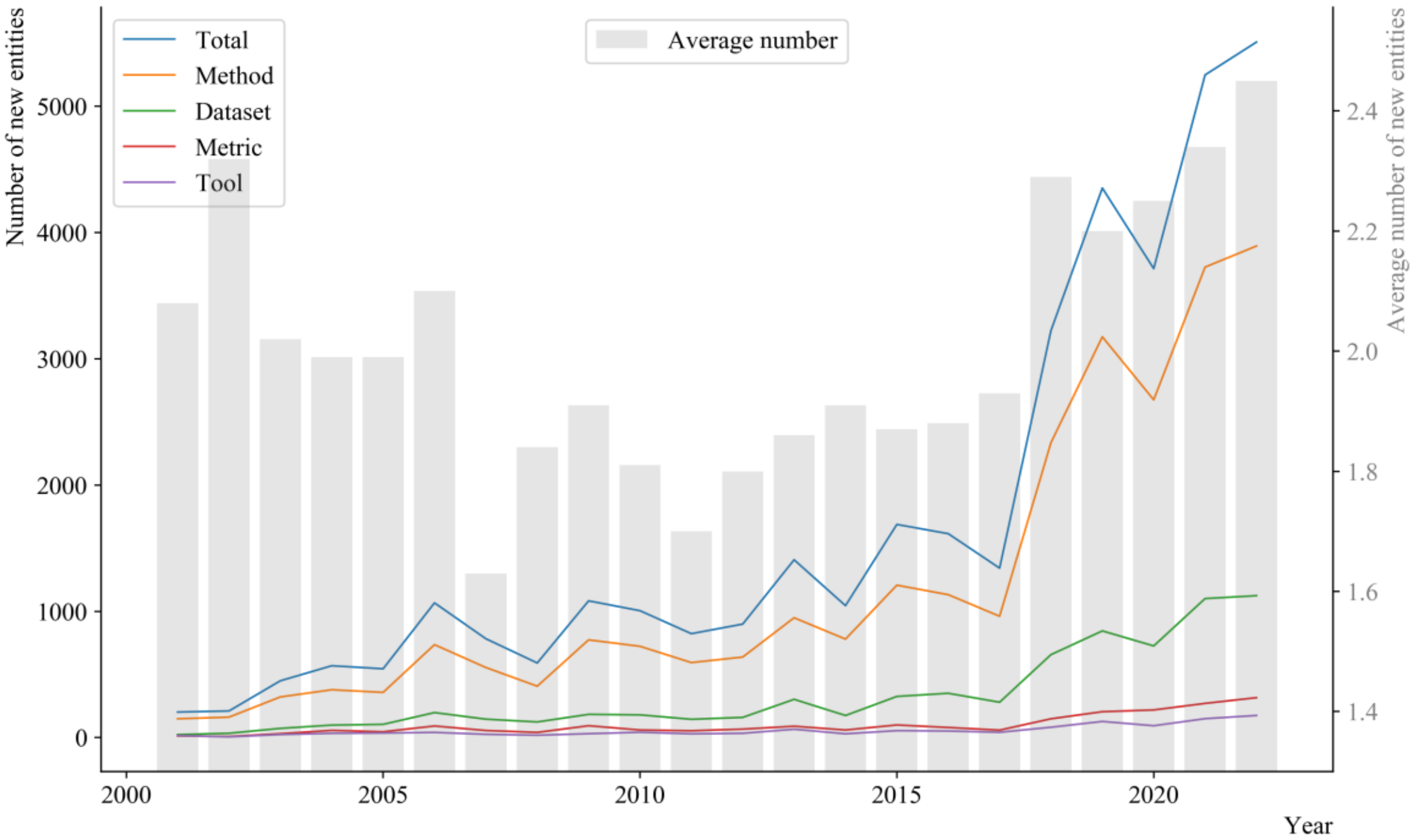

**Figure 6. Number of new technology-related entities each year**

It is worth pointing out that the number of new entities underwent a dramatic change in 2018. While the total number of new entities was 1342 in the previous year, it reached 3224 in 2018, representing a staggering growth rate of 140.2%. The number of new entities afterward far exceeded that before 2018. The significant increase in new entities from 2018 can be attributed to two main factors. Firstly, as demonstrated by Figure 2, the number of papers published in the NLP notably increased in 2018 compared to previous years. Additionally, large-scale pre-trained language models have undergone a surge in popularity since 2018, demonstrating substantial improvements over traditional machine learning models. In recent years, researchers have constructed a variety of network structures based on pre-trained models to tackle nearly all NLP tasks, including text classification, entity recognition, machine translation, automatic summarization, and more. This has resulted in the creation of many new method entities. Simultaneously, pre-trained models rely on a substantial amount of annotated data to fine-tune the models and achieve optimal results, leading to significant growth in the number of new dataset entities as well.

Let us examine the changes in the average number of new entities, with an overall trend of initially declining and then rising. During the initial years from 2001, the average number of new entities was higher than some of the intervening years. This could be attributed to the fact that we only collected entities that emerged in the field of NLP from research papers starting in the year 2000, thus missing out on many entities that had already emerged before 2000. Therefore, these previously omitted entities might have appeared in the years following 2001 and were counted as new entities for those years. Between 2007 and 2017, the average number of new entities remained at a low level, not exceeding 2. However, starting from 2018, the average number of new entities increased significantly compared to previous years. As mentioned earlier, there was a significant increase in both the number of papers and the number of new entities in 2018. It is well-known that pre-trained language models emerged and were introduced into the field of NLP around 2018. By examining the average number of new entities, we can to some extent control the impact of the growth in the papers number and confirm that pre-trained language models have injected new vitality into the technological innovation of the NLP field.

**(3) Technological innovation contributed by the majority of researchers**

To investigate which papers generated new technology-related entities, we calculated the percentage of documents that included at least one new entity annually starting from 2001. As illustrated in Figure 7, over 86% of the papers published each year contain at least one new technology-related entities. This finding suggests that most articles contribute to technological advancements in the field of NLP, indicating that the majority of researchers are actively involved in technological innovation. Perhaps the emphasis on technologies in the domain of NLP has led to this result.

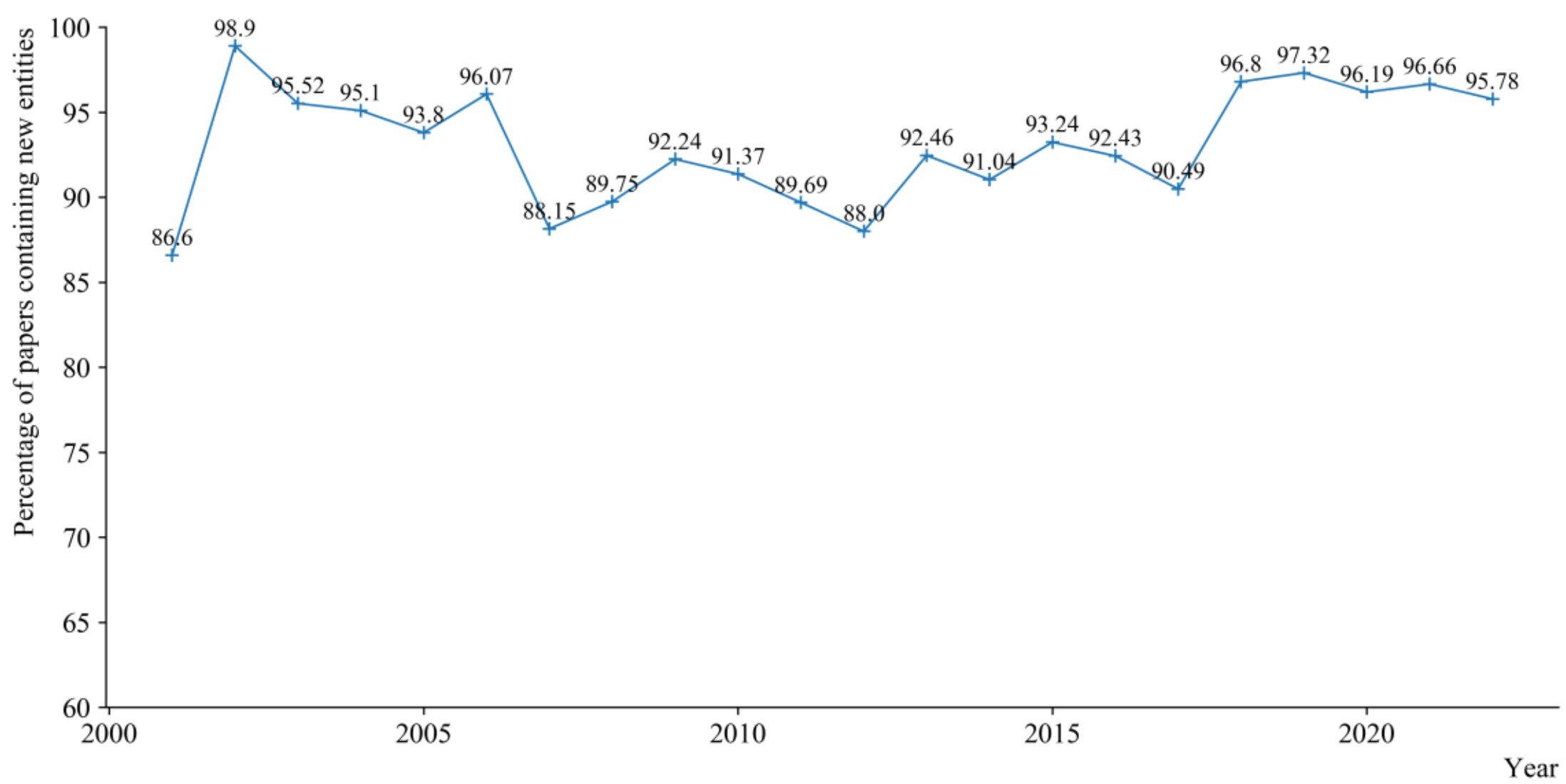

**Figure 7. Percentage of papers containing at least one new entity in each year**

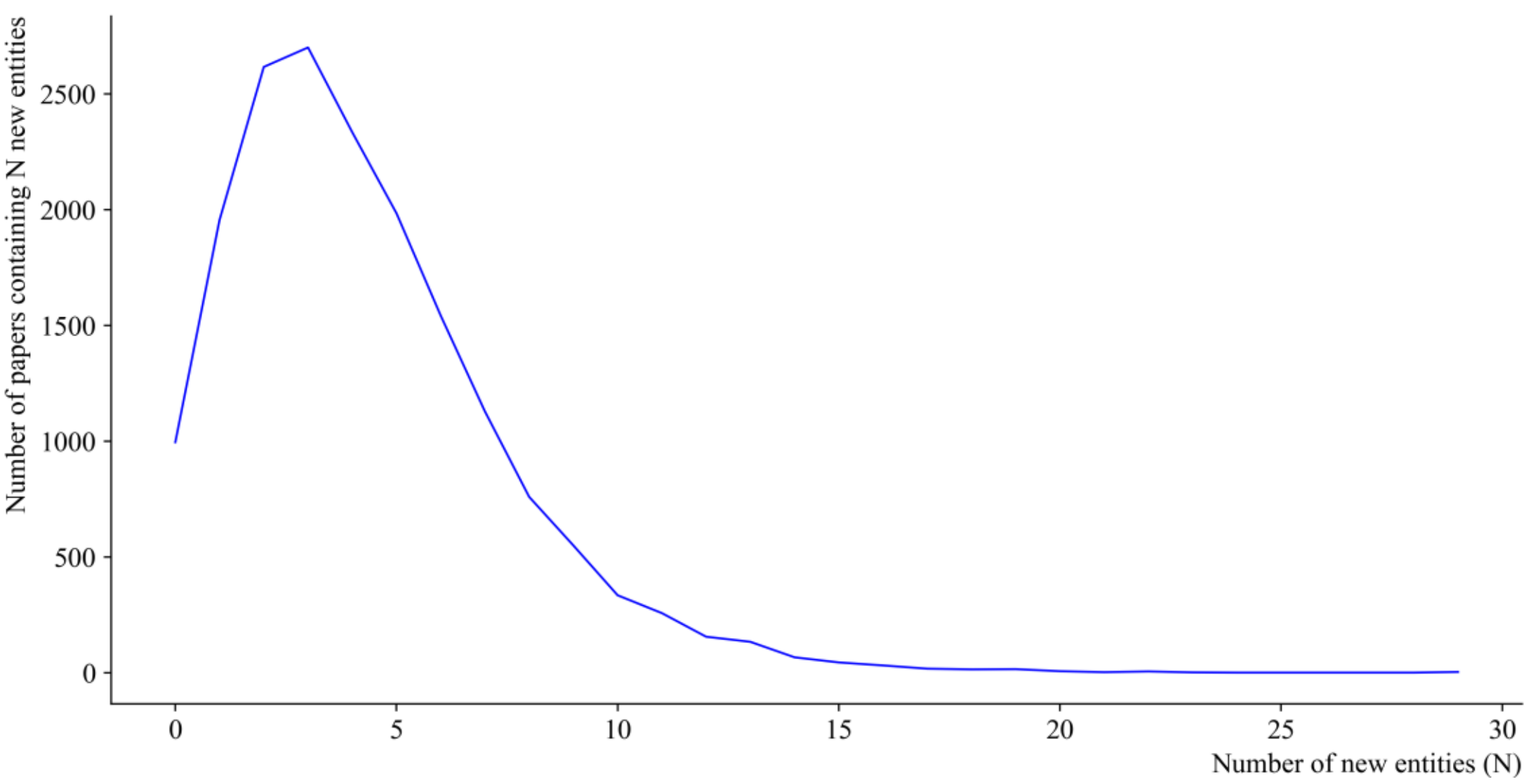

**Figure 8. Number distribution of papers containing *N* new entities**

To further explore the extent to which different papers contribute to technological innovation in NLP, we simplified the contribution of articles to technological innovation as the number of new technology-related entities included and counted the number of papers containing *N* new entities, the distribution of which is displayed in Figure 8. Only 996 papers do not mention any new entities, while 90% of the papers contain 1~10 new entities. After the number of new entities exceeding 15, there are very few corresponding papers. A single article contains up to 29 new entities, with only four such paper. There are many documents containing the number of new entities in the interval of 2~4, and the corresponding number of papers exceeds 2300 in each case. The articles containing three new entities are the most numerous, reaching 2701.

### 4.3 New technologies leading the development of NLP research

In this section, we answer RQ2 by identifying which technology-related entities are emerging and leading the development of the NLP domain since the 21st century. We calculate the impact of technology-related entities based on the co-occurrence networks of entities each year, and posit that high-impact technology-related entities are driving the development of NLP research. We further analyze the changes in the impact of these entities each year since their appearance.

**(1) High-impact entities of various types**

The normalized entities mentioned in at least three papers were selected to build network. We constructed co-occurrence networks of entities each year based on entity co-mentions in articles. And we calculated the *z-score* of technology-related entities in the networks to measure their impact. Guimerà et al. considered nodes with a *z-score* greater than 2.5 as critical nodes in the network community. We referred to this standard and identified entities with a *z-score* above 2.5 as having a high impact. It was worth noting that we considered the highest *z-score* of each entity after its first appearance as its impact. We then counted the number of new entities with high impact that have emerged since 2001, as illustrated in Table 9. Out of the 179 high-impact entities, method entities are found to be the most common, whereas tool entities were the least common. Dataset and metric entities are slightly more numerous than tool entities.

**Table 9. Number of high-impact new entities in four types**

| Entity type | Number of high-impact entities |
|---|---|
| Method | 130 |
| Dataset | 24 |
| Metric | 19 |
| Tool | 6 |
| Total | 179 |

The technologies involved in 179 high-impact entities have led the development of the NLP domain since the 21st century. We list the top 5 entities for each type and their *z-scores* in Table 10.

**Table 10. Top 5 entities in four types and their *z-scores***

| Type | Entity | *z-score* | Type | Entity | *z-score* |
|---|---|---|---|---|---|
| Method | BERT | 43.31 | Metric | BLEU | 15.93 |
| | Transformer | 34.67 | | Cross-Entropy | 13.13 |
| | LSTM | 28.82 | | ROUGE | 7.89 |
| | Attention Mechanism | 26.26 | | Fluency | 6.90 |
| | Adam | 20.36 | | Standard Deviation | 6.18 |
| Dataset | Wikipedia | 17.42 | Tool | PyTorch | 6.16 |
| | MNLI | 6.72 | | MOSES | 5.33 |
| | SQuAD | 5.78 | | GIZA++ | 5.21 |
| | Twitter | 5.31 | | TensorFlow | 3.56 |
| | SST-2 | 5.26 | | Stanford Parser | 3.30 |

Firstly, the z-scores reveal that method entities have the greatest impact, followed by dataset and metric entities, and tool entities with the least impact. These findings align with the analysis in section 4.2, which shows that researchers in the NLP domain prioritize methodological innovation. Secondly, we focus on the top 5 entities in each category. The top 5 method entities are all related to neural networks or pre-trained models. These entities have proven to be highly impactful, surpassing traditional machine learning methods that were popular in previous years. Among the top 5 dataset entities, Wikipedia corpus ranks first in impact, a dataset containing many full-text Wikipedias and becoming the preferred dataset for training word vectors and pre-trained models. Some datasets are available for specific tasks, such as the SQuAD (Stanford Question Answering Dataset) for reading comprehension and the SST-2 (Stanford Sentiment Treebank) dataset for sentiment analysis tasks. Among the top 5 metric entities, BLEU (Bilingual Evaluation Understudy) is the most impactful metric entity used to evaluate the performance of machine translation. At the same time, machine translation has been one of the most popular NLP tasks in recent years. ROUGE (Recall-Oriented Understudy for Gisting Evaluation) is a commonly used text summary evaluation metric, which can also be used for machine translation tasks. Standard deviation is a commonly used metric in statistical analysis to measure the variation of experimental results. Among the top 5 tool entities, PyTorch and TensorFlow are three deep learning frameworks on which researchers develop neural network models. PyTorch is becoming the most popular deep learning framework for researchers with its good API design and ease of use. And as more and more open-source NLP projects on GitHub use PyTorch, it becomes increasingly popular. Other tools are used for data processing in NLP processes, such as MOSES and GIZA++ for word alignment of the bilingual corpus, Stanford Parser for syntactic analysis.

**(2) *z-score* trend of high-impact entities**

In order to further investigate, we compared the highest *z-scores* (representing their peak impact) of all new entities post-2000, and found the top 10 entities with the highest impact. Their annual *z-scores*, post they were introduced into NLP are depicted in Figure 9. It is noteworthy that almost all of these entities are linked to deep learning and pre-trained language models, with no entities associated with traditional machine learning. It indicates that traditional machine learning methods have gradually receded in prominence within the field of NLP. Nowadays, when these methods are mentioned in papers, they are typically presented as background information or used as baseline models.

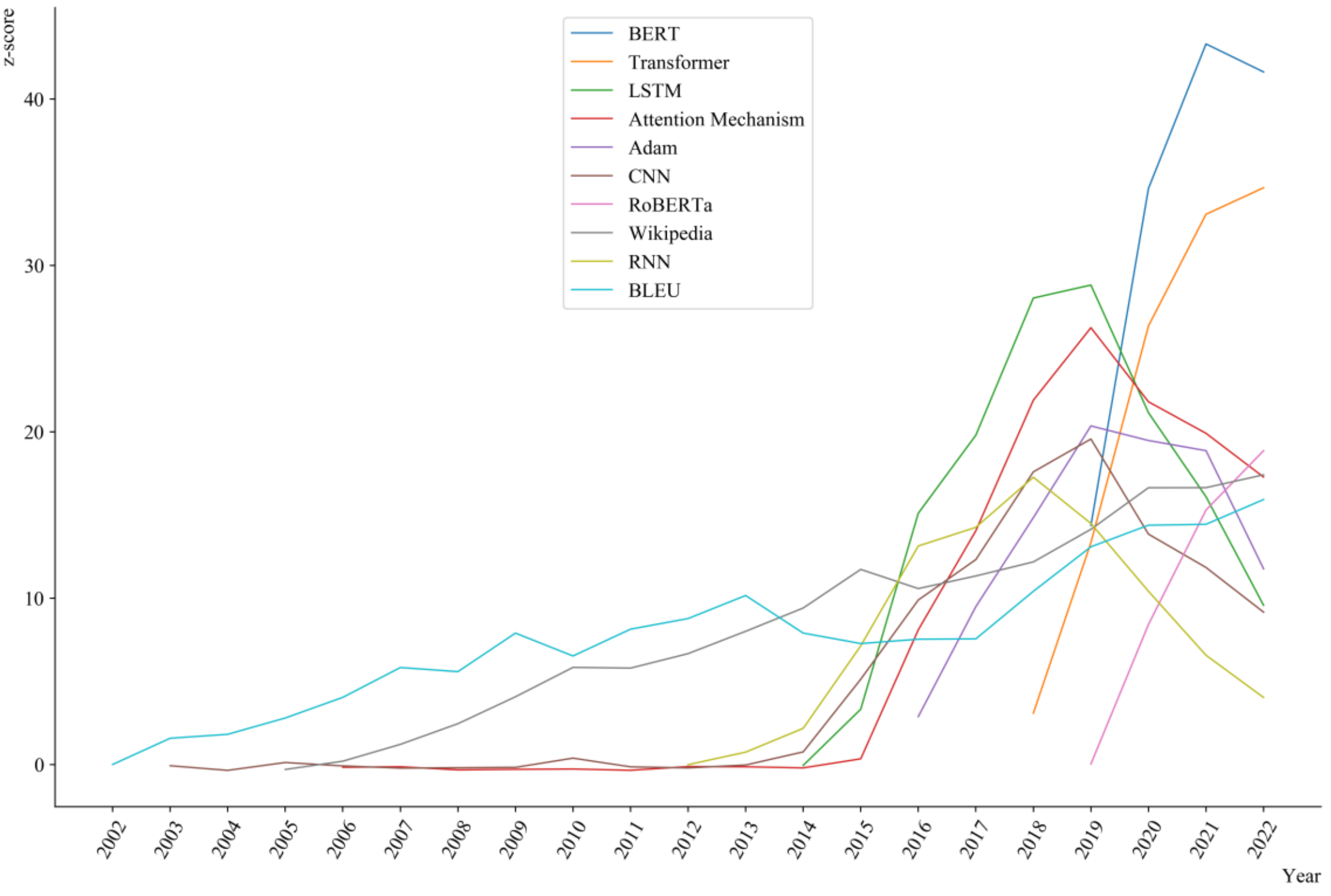

**Figure 9. *z-score* trend of top 10 new entities**

BERT (Bidirectional Encoder Representation from Transformers) has the highest impact, appearing in the NLP domain since 2019, which represents the beginning of pre-trained models used to address NLP problems. BERT achieved a high *z-score* in the year of its emergence. Then its impact increased yearly, reaching its highest *z-score* in 2021, surpassing all technology-related entities. And its impact slightly decreased in the year 2022. As an enhanced version of BERT, RoBERTa's (Robustly optimized BERT approach) z-score continues to rise year by year, but its influence still lags behind BERT significantly. The z-score of Transformer also maintains an upward trend year by year, and it has a significant impact. The entities related to deep learning are LSTM (Long Short-Term Memory), Attention mechanism, Adam (Adaptive Moment Estimation), CNN (Convolutional Neural Network), and RNN (Recursive Neural Network). They have the same trend of impact change. The *z-scores* of these five entities has been rising until 2018 or 2019 and downward since then. Before 2019, researchers used word vectors (e.g., GloVe) to construct embedding layers coupled with deep neural network structures (e.g., LSTM) to build models. This approach once replaced traditional machine learning methods as the mainstream of NLP studies. However, with the emergence of pre-trained language models, they outperform both traditional machine learning methods and general deep learning models on various NLP tasks. As a result, deep learning models such as LSTM have declined in impact after 2019.

The eight entities mentioned above are all methods, and the two remaining top 10 entities: are Wikipedia and BLEU (Bilingual Evaluation Understudy). One is a dataset, and the other is a metric. Their impact changes in a completely different way compared to the other eight entities, and overall, it has continued to show an upward trend. Wikipedia is a general corpus for word vector training and pre-trained language models. BLEU is a metric for evaluating machine translation, which has been one of the popular tasks in NLP. Therefore, the decline of general deep learning models and the uprising of pre-trained models did not affect the *z-score* of Wikipedia and BLEU.

### 4.4 The popularity of high-impact new technologies in recent years

We answer RQ3 in this section. According to the previous analysis, the number of new technology-related entities has increased significantly since 2018. We wanted to investigate whether the high-impact new technologies have become more popular and whether the popularity speed has accelerated over time. Concretely, we care about the changes in the cumulative *z-scores* of top new entities and the time the need to achieve high impact at each period.

#### (1) Cumulative *z-score* of the top 100 new entities in different periods

Since 2001, we have divided each period into three-year increments: 2001~2003, 2004~2006, 2007~2009, 2010~2012, 2013~2015, 2016~2018, and 2019~2021. During each period, we rank the *z-scores* of new entities separately for the year they appear and the following year. We accumulate the *z-scores* of the top 100 high-impact new entities. Specifically, we calculate the impact of the new entities that emerged in 2021 for the second year based on the entity co-occurrence network in 2022.

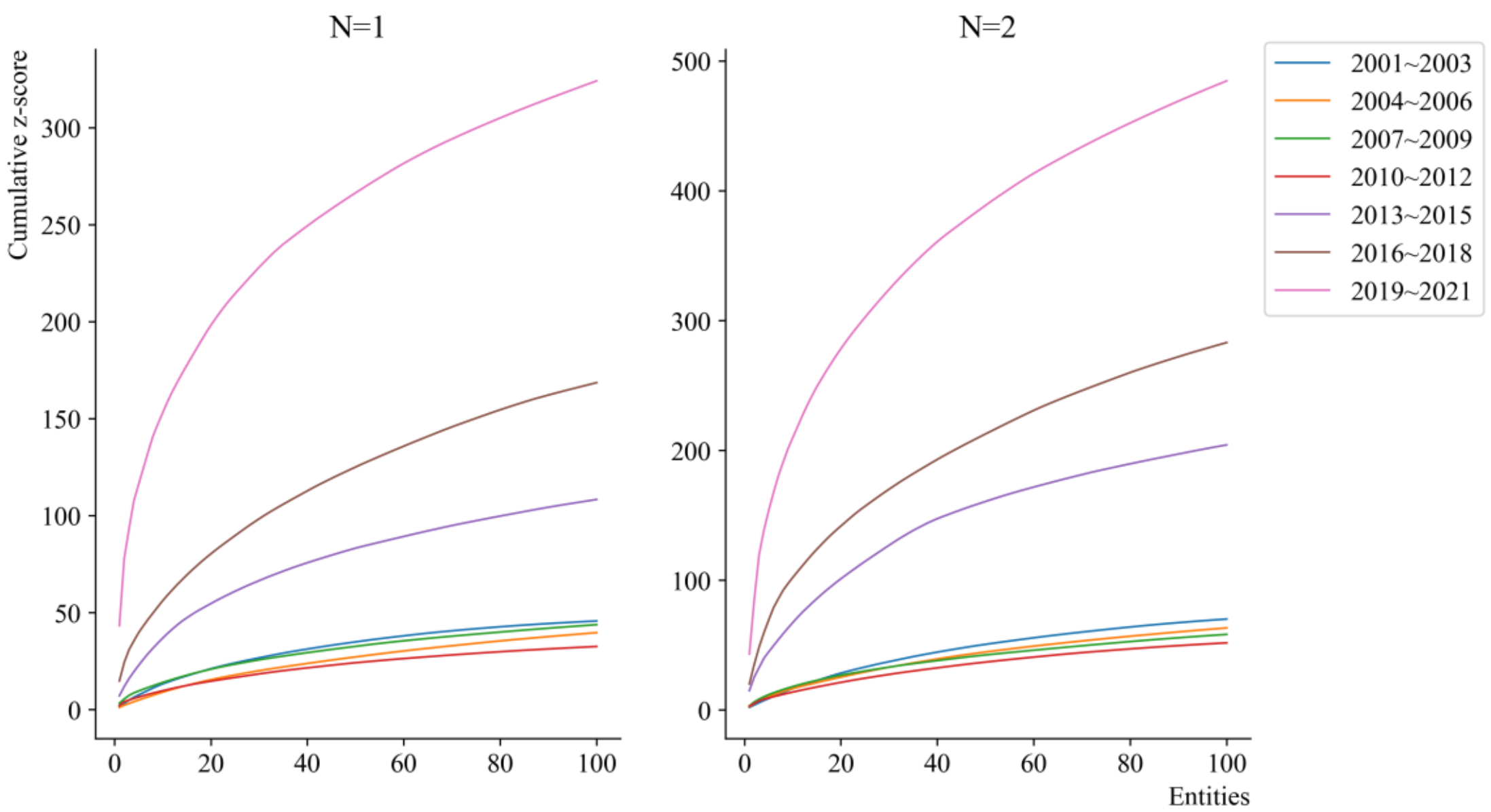

**Figure 10. Cumulative *z-score* of top new entities in the *N*-th year after their appearance**

As shown in Figure 10, the cumulative *z-score* of the top 100 new entities in 2010~2012 is the smallest. The cumulative *z-score* of the top 100 new entities during 2013~2015 showed a relatively significant increase compared to the earlier periods. But the cumulative *z-score* of the top 100 new entities from 2019 to 2021 is much higher than the other periods, regardless of *N*=1 or *N*=2. Around 2019, pre-trained language models were introduced into the field of NLP and have become well-received by researchers. They have quickly become the dominant research methods in NLP and are much more popular than traditional machine learning methods and general deep learning models. In recent years, the popularity of high-impact new technologies has far surpassed that of the past.

**(2) Average years of *z-score*>2.5 for high-impact new entities in different periods**

To further verify the conclusion about RQ3, we analyze the average time it takes for the high-impact new entity to have a *z-score* exceeding 2.5 in different periods. That is, whether the growth time required for new entities from emergence to achieving high level of impact is getting shorter. The analysis in section 4.3 shows that there are 179 new high-impact entities (*z-score*>2.5). We counted the number of high-impact new entities in each period as follows: 2001~2003: 32, 2004~2006: 28, 2007~2009: 13, 2010~2012: 11, 2013~2015: 33, 2016~2018: 37, and 2019~2021: 25. We then calculated the time from the appearance of these entities until their *z-scores* exceed 2.5. Appendix displays high-impact new entities from different periods and their corresponding highest *z-scores*. If the *z-score* of an entity is above 2.5 in the year it appears, then the time is one year. The average time for new entities to reach a *z-score*>2.5 for different periods is presented in Figure 11.

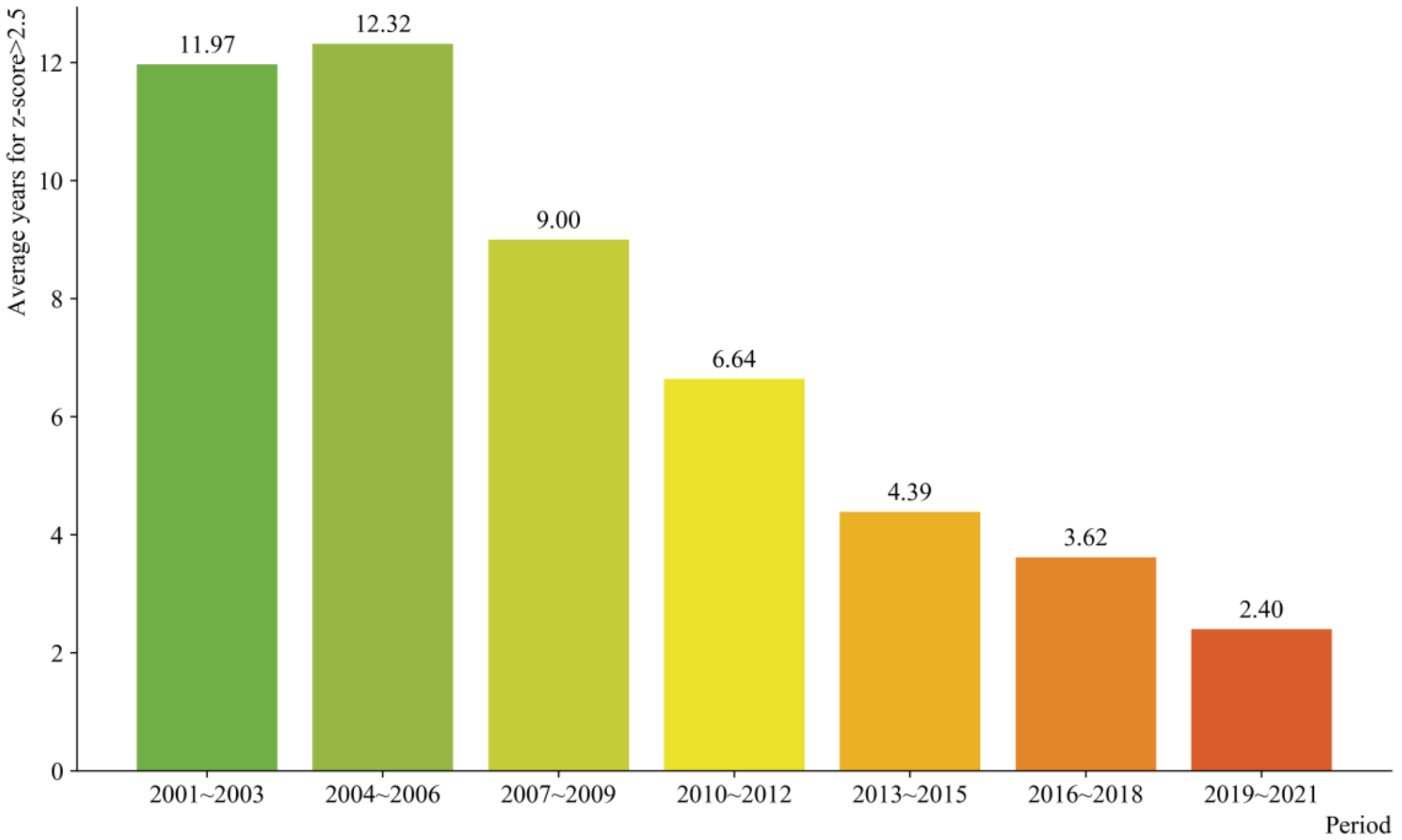

**Figure 11. Average years for new entities to reach *z-score*>2.5 in different periods**

As observed in Figure 11, since the 21st century, technology-related entities that appeared earlier in NLP have taken more time to become high-impact entities on average. In the periods of 2001~2003, 2004~2006, the average time was 12.03 years and 12.39 years, respectively. And then, the average time gradually declined, to an average of 3.62 years in 2016~2018. In 2019~2021, high-impact entities will only need an average of 2.40 years for their *z-scores* to exceed 2.5. Some technology-related entities achieved high impact in the year they were introduced to the NLP field. For example, BERT first appeared in the NLP domain in 2019 and got a *z-score* of 14.37 that year.

Pre-trained language models first appeared in computer science and were verified to perform far superior to previous models. As a result, they were introduced into the NLP field and became much more impactful than previous NLP techniques. Moreover, pre-trained language models quickly became mainstream research methods in NLP from their emergence. The excellent performance of pre-trained language models has brought a boom in NLP research in recent years. Meanwhile, we are looking forward to the next revolution of NLP technologies.

### 4.5 Comparison to existing studies conducted from the topic perspective

In this section, we compare our analysis of technology development from an entity-centric perspective with existing research on the development of the NLP field from the topic perspective. Parmar et al. (2020) summarized the major research topics in the field of NLP from 1965 to 2018. Hingmire et al. (2021) constructed a topic taxonomy based on a corpus of resources form NLP and related fields, whereby 23,293 resources were manually assigned nodes of the taxonomy. In our study, we conducted a detailed and precise entity analysis of the technological development in the field of NLP from 2000 to 2022, surpassing the granularity of traditional research topics. The research process relied significantly less on manual involvement, and our crucial processes of entity extraction and normalization have undergone rigorous evaluations. Leveraging well-trained models for entity recognition, alongside a suite of entity normalization and data processing procedures, we have the capability to automate and swiftly update the analysis for future data. In contrast, the aforementioned studies, due to their reliance on manual intervention for research topic formulation and classification, cannot achieve the same level of convenience for continuous updates.

Mohammad (2019) conducted a statistical analysis of bigrams in NLP conference paper titles, revealing that the occurrences of “neural machine” have been increasing since 2016, while the occurrences of “statistical machine” have been steadily declining since an earlier period. This conclusion is similar to the findings illustrated in Figure 9 of our research. However, by employing z-score calculations, we have provided a more compelling validation for the increasing impact of specific pre-training models such as “BERT”, “RoBERTa”, and “Transformer” in the field of NLP since 2018, coupled with a diminishing influence of neural network architectures like “LSTM” and “CNN”. Furthermore, the appendix contains a comprehensive listing of high-impact technology entities in different time periods. It is worth noting that relying solely on bigrams extracted from paper titles is a crude approach, which can only offer a rough indication of the research topics. In contrast, our methodology involves the extraction and normalization of technology-related entities derived from pre-trained models, resulting in more accurate representations. Building upon this foundation, we conducted a quantitative analysis of these entities within the full text of the articles and derived valuable insights, which are elaborated in sections 4.2, 4.3, and 4.4.

## 5. Discussion

In this paper, our main objective is to analyze the development of technology in the NLP domain. To achieve this goal, we have developed an automatic scientific entity identification model that is combined with semi-supervised data augmentation to extract method, dataset, metric, and tool entities from NLP articles. Furthermore, a semi-automatic approach has been devised for normalizing the extracted entities. We then constructed entity co-occurrence networks based on each year's paper data and calculated the impact of entities in each year. After the above preparation, we investigate the development of NLP technologies from the entity level.

### 5.1 Implications

We elaborate on the implications of this paper from both theoretical and practical perspectives.

**(1) Theoretical implications**

This study provides a new perspective and methodology for analyzing the development of technologies in a specific domain through fine-grained entity analysis. Previous studies on domain development have typically utilized topic models to examine changes at a coarse level. Our approach demonstrates the feasibility of gaining more nuanced insights by focusing on technology-related entities.

This study showcases the effectiveness of automatic scientific entity recognition and network analysis in uncovering the patterns of technology development within a domain. This enables the potential generalization of such methods to other fields. The proposed methods, including entity recognition model, data augmentation technique, entity normalization, and impact measurement based on co-occurrence networks, are not limited to the NLP field.

The findings reveal the growing complexity of research in the NLP domain as reflected by the increasing number of technology entities within each paper. This provides a novel perspective for comprehending the knowledge burden placed on researchers and establishes basis for corresponding educational and support efforts.

**(2) Practical implications**

In this article, we have identified high-impact technologies that have driven NLP advancements over the past two decades, providing valuable insights for researchers and practitioners to prioritize learning and applying these important innovations.

The analysis of shortening time periods for new technologies to gain popularity implies their rapid adoption, posing new demands for NLP practitioners to update skills.

This study provides evidence of the democratization of technology innovation with most papers contributing new entities. This points to the importance of encouraging broad participation.

Plotting trajectories of impact changes for specific technologies helps understand their rise and fall patterns. These trends can guide strategic technology investment and staffing decisions in industry.

In summary, the entity-level technological development analysis offers new insights for both theoretical research and practical applications. This study demonstrates the value of such an approach and lays the foundation for future work.

**5.2 Limitations**

Indeed, there are some limitations in this study. The accuracy of our automatic recognition of scientific entities in academic papers still needs to be improved and is not yet comparable to entity recognition in common domains. Additionally, the precision of the semi-automatic entity normalization method that we have developed still has room for improvement, and there is a need to reduce human involvement in the normalization process even further. Both incorrectly identified entities and errors arising in normalization can affect later calculations of entity impact. However, we analyze overall development trends or conducted case studies based on high-impact technology-related entities. The error in the calculation processes should have a minor effect on the comprehensive analysis and the high-frequency entities, thus making our conclusions reliable. Furthermore, we compared the scientific entity recognition model we created with models from other researchers to demonstrate its superior performance. Finally, all research data in this paper were obtained from the proceedings of three conferences, ACL, EMNLP, and NAACL, from 2000 to 2022. Although these three conferences are the major conferences in NLP, using their data to represent the entire domain is necessarily not comprehensive enough. We did not collect data on other conference and journal articles in the field of NLP, nor did we collect data before 2000.

## 6. Conclusion and future works

In this article, we proposed to analyze the technological development in a specific domain (NLP) from an entity perspective, to gain more nuanced insights beyond the coarse-grained topic level. Initially, we built a scientific entity recognition model based on existing works, and combined it with semi-supervised data augmentation to identify the technology-related entities in NLP papers. Subsequently, a semi-automatic approach was devised to normalize the extracted entities, followed by the construction of co-occurrence networks based on the co-mention of entities in papers every year, thereby enabling the computation of entities' impact. After completing the preparations, we formulated three research questions to analyze the quantitative alterations of technology-related entities in the NLP field, the impact changes of top entities, as well as the popularity degree and acceptance speed of high-impact technology-related entities.

In future works, firstly, we should strive to improve the accuracy of scientific entity recognition and normalization. Secondly, the number of new technology-related entities has significantly exceeded that of the past since 2018, further clustering analysis for these entities can be conducted. Thirdly, researchers generally use a combination of technologies to solve problems, so we will consider identifying combinations of technologies in co-occurrence networks and analyzing the development of "technology communities". Furthermore, we can employ clustering methods to study the evolutionary pathways between them. Moreover, to accurately observe the usage of technologies, we should combine them with the research questions of articles. We can then analyze how researchers have changed the combinations of technologies they use over time for the same research topic. This analysis can identify which research topics have only undergone slight changes in technology combinations and which are constantly evolving.

**Acknowledgments**

This paper was supported by the National Natural Science Foundation of China (Grant No.72074113, 72374103).

**Appendix**

**High-impact new entities of different periods and their highest *z-scores***

| Periods | Entities and *z-score* |
| --- | --- |
| 2001~2003 | beam search: 9.29, bag-of-words: 7.92, reinforcement learning: 7.89, SMT: 6.89, PCFG: 3.34, supervised learning: 3.03, NN: 2.68, learning algorithm: 2.53, BLEU: 15.93, TF-IDF: 6.67, log-linear model: 5.04, NER: 4.08, PMI: 3.49, runtime: 2.87, perceptron: 2.84, Hits @ 1: 2.74, LDC: 2.56, MRR: 2.52, CNN: 19.57, cosine similarity: 6.57, CRF: 6.25, standard deviation: 6.18, GIZA++: 5.21, KL divergence: 4.24, train: 4.09, phrase-based system: 3.78, n-gram language model: 3.73, BM25: 3.54, t-test: 3.14, CoNLL: 3.02, Kneser-Ney smoothing: 2.84, coherence: 2.74. |
| 2004~2006 | MLP: 8.52, ROUGE: 7.89, fluency: 6.90, ROUGE-L: 5.72, logistic regression: 5.49, graph: 4.24, linear model: 3.81, MT: 3.11, Performance: 2.85, CoNLL shared tasks: 2.63, Wikipedia: 17.42, cross-entropy: 13.13, MERT: 5.54, NLP: 4.02, QA: 3.75, RL: 3.74, task: 3.58, ROUGE-1: 3.49, RTE: 2.99, ROUGE-2: 2.93, PBSMT: 2.75, MSE: 2.74, Attention Mechanism: 26.26, Gigaword: 4.47, METEOR: 4.20, LDA: 4.09, Stanford parser: 3.30, mean: 2.65. |
| 2007~2009 | PLMs: 8.79, SGD: 7.77, Moses: 5.33, TM: 4.13, informativeness: 2.81, multi-task learning: 6.96, transfer learning: 6.23, Amazon Mechanical Turk: 4.70, English Wikipedia: 3.98, SGD: 3.67, span: 2.79, SST-2: 5.26, Freebase: 3.88 |
| 2010~2012 | backpropagation: 6.12, Twitter: 5.31, WMT: 2.99, grid search: 2.98, crowdsourcing: 2.64, softmax: 10.57, GLUE: 3.69, recursive neural network: 17.27, recurrent neural network: 10.84, deep learning: 8.44, stochastic gradient descent (SGD): 2.92. |
| 2013~2015 | recurrent neural network (RNN): 7.98, FFN: 7.41, deep neural networks: 6.97, convolutional: 5.86, MT: 4.58, feed-forward neural network: 4.57, AdaGrad: 4.24, Reddit: 3.27, tanh: 3.03, recurrent networks: 2.52, LSTM: 28.82, GloVe: 12.77, word2vec: 10.40, skip-gram: 5.00, NLI: 4.58, max-pooling: 4.19, dropout: 4.07, t-SNE: 3.64, CBOW: 3.17, KG: 3.01, ReLU: 3.00, bidirectional RNN: 2.95, end-to-end: 2.95, validation: 2.71, BiLSTM: 13.27, Seq2Seq: 13.12, GRU: 8.43, contrastive learning: 6.97, NMT: 6.22, Adadelta: 3.92, MRPC: 3.23, SNLI: 2.83, Theano: 2.55. |
| 2016~2018 | Adam: 20.36, self-attention: 12.48, encoder-decoder model: 6.90, SQuAD: 5.78, BPE: 5.36, copy mechanism: 4.49, TensorFlow: 3.56, KD: 3.30, GloVe embeddings: 3.24, attention weights: 3.15, VAE: 3.02, Wikidata: 2.84, BiGRU: 2.64, teacher model: 2.56, crowdworkers: 2.56, sequence-to-sequence (seq2seq): 2.53, PyTorch: 6.16, fastText: 5.09, cross-attention: 5.06, GCN: 4.58, adversarial training: 3.97, back-translation: 2.87, adversarial: 2.61, Transformer: 34.67, ELMo: 9.88, Transformer encoder: 7.12, MNLI: 6.72, GNN: 5.16, multi-head attention: 4.72, TLM: 4.66, attention heads: 3.64, CLS: 3.36, QQP: 3.04, few-shot learning: 2.75, selfattention: 2.70, STS-B: 2.68, autoregressive models: 2.51. |
| 2019~2021 | BERT: 43.31, RoBERTa: 18.87, GPT-2: 10.66, MLM: 10.31, HuggingFace: 6.20, BERTScore: 5.43, Phrase-BERT: 5.19, GPT: 4.74, mBERT: 4.66, XLNet: 4.36, BERT encoder: 4.26, bert-base-uncased: 4.17, SBERT: 3.68, QNLI: 2.77, encoder-decoder Transformer: 2.67, CoLA: 2.58, fairseq: 2.53, T5: 13.84, BART: 12.81, GPT-3: 7.33, AdamW: 6.97, XLM-R: 5.00, XPROMPT: 3.68, DistilBERT: 2.80, ALBERT: 2.71. |

**References**

Beltagy, I., Lo, K., & Cohan, A. (2019). SciBERT: A Pretrained Language Model for Scientific Text. *Proceedings of the 2019 Conference on Empirical Methods in Natural Language Processing and the 9th International Joint Conference on Natural Language Processing (EMNLP-IJCNLP)*, 3615–3620. https://doi.org/10.18653/v1/D19-1371

Brooks, H. (1980). Technology, Evolution, and Purpose. *Daedalus*, *109*(1), 65–81. JSTOR.

Chalkidis, I., Fergadiotis, M., Malakasiotis, P., Aletras, N., & Androutsopoulos, I. (2020). LEGAL-BERT: The Muppets straight out of Law School. *Findings of the Association for Computational Linguistics: EMNLP 2020*, 2898–2904. https://doi.org/10.18653/v1/2020.findings-emnlp.261

Coccia, M. (2019). The theory of technological parasitism for the measurement of the evolution of technology and technological forecasting. *Technological Forecasting and Social Change*, *141*, 289–304. https://doi.org/10.1016/j.techfore.2018.12.012

Ding, Y., Song, M., Han, J., Yu, Q., Yan, E., Lin, L., & Chambers, T. (2013). Entitymetrics: Measuring the Impact of Entities. *PLOS ONE*, *8*(8), e71416. https://doi.org/10.1371/journal.pone.0071416

Dzau, V. J., & Balatbat, C. A. (2018). Health and societal implications of medical and technological advances. *Science Translational Medicine*, *10*(463), eaau4778. https://doi.org/10.1126/scitranslmed.aau4778

Eberts, M., & Ulges, A. (2021). *Span-based Joint Entity and Relation Extraction with Transformer Pre-training* (arXiv:1909.07755). arXiv. https://doi.org/10.3233/FAIA200321

Funk, R. J., & Owen-Smith, J. (2017). A Dynamic Network Measure of Technological Change. *Management Science*, *63*(3), 791–817. https://doi.org/10.1287/mnsc.2015.2366

Geng, R., Chen, Y., Huang, R., Qin, Y., & Zheng, Q. (2023). Planarized sentence representation for nested named entity recognition. *Information Processing & Management*, *60*(4), 103352. https://doi.org/10.1016/j.ipm.2023.103352

Guimerà, R., Sales-Pardo, M., & Amaral, L. A. N. (2007). Classes of complex networks defined by role-to-role connectivity profiles. *Nature Physics*, *3*(1), Article 1. https://doi.org/10.1038/nphys489

Hingmire, S., Li, I., Kawamura, R., Chen, B., Fabbri, A., Tang, X., Liu, Y., George, T., Liao, T., Wong, W. P., Yan, V., Zhou, R., Palshikar, G. K., & Radev, D. (2021). *CLICKER: A Computational LInguistics Classification Scheme for Educational Resources* (arXiv:2112.08578). arXiv. http://arxiv.org/abs/2112.08578

Hou, Y., Jochim, C., Gleize, M., Bonin, F., & Ganguly, D. (2021). TDMSci: A Specialized Corpus for Scientific Literature Entity Tagging of Tasks Datasets and Metrics. *Proceedings of the 16th Conference of the European Chapter of the Association for Computational Linguistics: Main Volume*, 707–714. https://doi.org/10.18653/v1/2021.eacl-main.59

Hu, Z. (2022). Study of the Effectiveness of 5G Mobile Internet Technology to Promote the Reform of English Teaching in the Universities and Colleges. *Computational Intelligence and Neuroscience*,

*2022*, 1–8. https://doi.org/10.1155/2022/3053694

Huang, L., Chen, X., Zhang, Y., Wang, C., Cao, X., & Liu, J. (2022). Identification of topic evolution: Network analytics with piecewise linear representation and word embedding. *Scientometrics*, *127*(9), 5353–5383. https://doi.org/10.1007/s11192-022-04273-1

Järvelin, K., & Vakkari, P. (1993). The evolution of library and information science 1965–1985: A content analysis of journal articles. *Information Processing & Management*, *29*(1), 129–144. https://doi.org/10.1016/0306-4573(93)90028-C

Khan, N., Ray, R. L., Kassem, H. S., & Zhang, S. (2022). Mobile Internet Technology Adoption for Sustainable Agriculture: Evidence from Wheat Farmers. *Applied Sciences*, *12*(10), 4902. https://doi.org/10.3390/app12104902

Lee, J., Yoon, W., Kim, S., Kim, D., Kim, S., So, C. H., & Kang, J. (2020). BioBERT: A pre-trained biomedical language representation model for biomedical text mining. *Bioinformatics*, *36*(4), 1234–1240. https://doi.org/10.1093/bioinformatics/btz682

Li, H., An, H., Wang, Y., Huang, J., & Gao, X. (2016). Evolutionary features of academic articles co-keyword network and keywords co-occurrence network: Based on two-mode affiliation network. *Physica A: Statistical Mechanics and Its Applications*, *450*, 657–669. https://doi.org/10.1016/j.physa.2016.01.017

Li, K., & Yan, E. (2018). Co-mention network of R packages: Scientific impact and clustering structure. *Journal of Informetrics*, *12*(1), 87–100. https://doi.org/10.1016/j.joi.2017.12.001

Liu, T., & Liu, M. (2022). Effect of Mobile Internet Technology in Health Management of Heart Failure Patients Guiding Cardiac Rehabilitation. *Journal of Healthcare Engineering*, *2022*, 1–5. https://doi.org/10.1155/2022/7118919

Liu, Z., Huang, D., Huang, K., Li, Z., & Zhao, J. (2020). FinBERT: A Pre-trained Financial Language Representation Model for Financial Text Mining. *Proceedings of the Twenty-Ninth International Joint Conference on Artificial Intelligence*, 4513–4519. https://doi.org/10.24963/ijcai.2020/622

Luan, Y., He, L., Ostendorf, M., & Hajishirzi, H. (2018). Multi-Task Identification of Entities, Relations, and Coreference for Scientific Knowledge Graph Construction. *Proceedings of the 2018 Conference on Empirical Methods in Natural Language Processing*, 3219–3232. https://doi.org/10.18653/v1/D18-1360

Ma, Y., Liu, J., Lu, W., & Cheng, Q. (2023). From "what" to "how": Extracting the Procedural Scientific Information Toward the Metric-optimization in AI. *Information Processing & Management*, *60*(3), 103315. https://doi.org/10.1016/j.ipm.2023.103315

Mao, X., Huang, S., Li, R., & Shen, L. (2020). Automatic Keywords Extraction Based on Co-Occurrence and Semantic Relationships Between Words. *IEEE Access*, *8*, 117528–117538. https://doi.org/10.1109/ACCESS.2020.3004628

McHugh, M. L. (2012). Interrater reliability: The kappa statistic. *Biochemia Medica*, *22*(3), 276–282. https://doi.org/10.11613/BM.2012.031

Mitcham, C. (1994). *Thinking through technology: The path between engineering and philosophy*. University of Chicago Press.

Mohammad, S. M. (2019). *The State of NLP Literature: A Diachronic Analysis of the ACL Anthology* (arXiv:1911.03562). arXiv. http://arxiv.org/abs/1911.03562

Parmar, M., Jain, N., Jain, P., Jayakrishna Sahit, P., Pachpande, S., Singh, S., & Singh, M. (2020). NLPExplorer: Exploring the Universe of NLP Papers. In J. M. Jose, E. Yilmaz, J. Magalhães, P. Castells, N. Ferro, M. J. Silva, & F. Martins (Eds.), *Advances in Information Retrieval* (pp. 476–

480). Springer International Publishing. https://doi.org/10.1007/978-3-030-45442-5_61

Percia David, D., Maréchal, L., Lacube, W., Gillard, S., Tsesmelis, M., Maillart, T., & Mermoud, A. (2023). Measuring security development in information technologies: A scientometric framework using arXiv e-prints. *Technological Forecasting and Social Change*, *188*, 122316. https://doi.org/10.1016/j.techfore.2023.122316

Rotolo, D., Hicks, D., & Martin, B. R. (2015). What is an emerging technology? *Research Policy*, *44*(10), 1827–1843. https://doi.org/10.1016/j.respol.2015.06.006

Sigman, M. (2020). Telemedicine in reproductive medicine—Implications for technology and clinical practice. *Fertility and Sterility*, *114*(6), 1125. https://doi.org/10.1016/j.fertnstert.2020.10.042

Skolnikoff, E. B. (1993). *The Elusive Transformation: Science, Technology, and the Evolution of International Politics*. Princeton University Press. https://www.jstor.org/stable/j.ctt7rpm1

Sundheim, B. M. (1995). Named entity task definition, version 2.1. *Proceedings of the Sixth Message Understanding Conference (MUC-6)*.

Tuomaala, O., Järvelin, K., & Vakkari, P. (2014). Evolution of Library and Information Science, 1965-2005: Content Analysis of Journal Articles. *Journal of the Association for Information Science and Technology*, *65*(7), 1446–1462. https://doi.org/10.1002/asi.23034

Wang, X., Cheng, Q., & Lu, W. (2014). Analyzing evolution of research topics with NEViewer: A new method based on dynamic co-word networks. *Scientometrics*, *101*(2), 1253–1271. https://doi.org/10.1007/s11192-014-1347-y

Wei, Z., Hong, Y., Zou, B., Cheng, M., & Yao, J. (2020). Don't Eclipse Your Arts Due to Small Discrepancies: Boundary Repositioning with a Pointer Network for Aspect Extraction. *Proceedings of the 58th Annual Meeting of the Association for Computational Linguistics*, 3678–3684.

https://doi.org/10.18653/v1/2020.acl-main.339

Wei, Z., Su, J., Wang, Y., Tian, Y., & Chang, Y. (2020). A Novel Cascade Binary Tagging Framework for Relational Triple Extraction. *Proceedings of the 58th Annual Meeting of the Association for Computational Linguistics*, 1476–1488. https://doi.org/10.18653/v1/2020.acl-main.136

Yao, R., Ye, Y., Zhang, J., Li, S., & Wu, O. (2023). Exploring developments of the AI field from the perspective of methods, datasets, and metrics. *Information Processing & Management*, *60*(2), 103157. https://doi.org/10.1016/j.ipm.2022.103157

Yun, S., Cho, W., Kim, C., & Lee, S. (2022). Technological trend mining: Identifying new technology opportunities using patent semantic analysis. *Information Processing & Management*, *59*(4), 102993. https://doi.org/10.1016/j.ipm.2022.102993

Zaratiana, U., Holat, P., Tomeh, N., & Charnois, T. (2022). *Hierarchical Transformer Model for Scientific Named Entity Recognition* (arXiv:2203.14710). arXiv. http://arxiv.org/abs/2203.14710

Zhang, C., Mayr, P., Lu, W., & Zhang, Y. (2023). Guest editorial: Extraction and evaluation of knowledge entities in the age of artificial intelligence. *Aslib Journal of Information Management*, *75*(3), 433–437. https://doi.org/10.1108/AJIM-05-2023-507

Zhang, Y., Zhang, G., Zhu, D., & Lu, J. (2017). Scientific evolutionary pathways: Identifying and visualizing relationships for scientific topics. *Journal of the Association for Information Science and Technology*, *68*(8), 1925–1939. https://doi.org/10.1002/asi.23814

Zhong, Z., & Chen, D. (2021). A Frustratingly Easy Approach for Entity and Relation Extraction. *Proceedings of the 2021 Conference of the North American Chapter of the Association for Computational Linguistics: Human Language Technologies*, 50–61. https://doi.org/10.18653/v1/2021.naacl-main.5